\documentclass[runningheads]{llncs}
\usepackage[T1]{fontenc}
\usepackage{graphicx}
\usepackage{amsmath} % For 'split' environment
\usepackage{amssymb} % For '\mathbb' command
\usepackage{xcolor}
\begin{document}
\title{Fine-grained Text to Image Synthesis}
\author{Xu Ouyang \and
Ying Chen \and
Kaiyue Zhu \and
Gady Agam}
\authorrunning{X. Ouyang et al.}

\institute{Illinois Institute of Technology,
Chicago, Illinois, USA\\
\email{\{xouyang3, ychen245, kzhu6\}@hawk.iit.edu, agam@iit.edu}}
\maketitle              % typeset the header of the contribution
\begin{abstract}
 Fine-grained text to image synthesis involves generating images from texts that belong to different categories. In contrast to general text to image synthesis, in fine-grained synthesis there is high similarity between images of different subclasses, and there may be linguistic discrepancy among texts describing the same image. Recent Generative Adversarial Networks (GAN), such as the Recurrent Affine Transformation (RAT) GAN model, are able to synthesize clear and realistic images from texts. However, GAN models ignore fine-grained level information. In this paper we propose an approach that incorporates an auxiliary classifier in the discriminator and a contrastive learning method to improve the accuracy of fine-grained details in images synthesized by RAT GAN. The auxiliary classifier helps the discriminator classify the class of images, and helps the generator synthesize more accurate fine-grained images. The contrastive learning method minimizes the similarity between images from different subclasses and maximizes the similarity between images from the same subclass. We evaluate on several state-of-the-art methods on the commonly used CUB-200-2011 bird dataset and Oxford-102 flower dataset, and demonstrated superior performance.

\end{abstract}
\section{Introduction}
Text to image synthesis is a fundamental problem due to gaps between text with limited information and high-resolution image with rich contents. Currently, there are three main approaches to solve this problem. The first approach is based on Generative Adversarial Networks (GANs)~\cite{NIPS2014Goodfellow} and have achieved great success in image synthesis. GANs involves two neural networks that work in opposition as a zero-sum game: a generator that synthesizes fake image and a discriminator that evaluates whether images are fake or real. GAN approaches to synthesis include: Conditional GAN for synthesizing an image from sentence-level text, LSTM conditional GAN~\cite{ICPR2018Xu} for synthesizing images from word-level text, and fine-grained text to image synthesis based on attention~\cite{CVPR2018Tao}. Language-free text to image synthesis (LAFITE)~\cite{CVPR2022Yufan} was proposed based on the Stylegan2 and CLIP models. Text and image fusion during image synthesis using a recurrent affine transformation (RAT) GAN model was proposed in~\cite{Arxiv2022Senmao}. All of these approaches focus on generating high-quality images, but neglect the differences between subclasses within the dataset. This can result in varying degrees of similarity among synthesized images from different subclasses and negatively affect performance.

The second approach for text to image synthesis is based on Auto Regressive Generative models, which treat text to image synthesis as a transformation from textual tokens to visual tokens based on a sequence-to-sequence Transformer model. DALL-E~\cite{ICML2021Aditya} and CogView~\cite{NIPS2021Ming} both aim to learn the relationship between texts and images based on a Transformer model. They first convert the image into a sequence of discrete image tokens with Vector Quantized Variational Autoencoder (VQ-VAE)~\cite{NIPS2017Aaron}, and then convert text tokens into image tokens by using a sequence-to-sequence Transformer, as both text and image are formatted as sequences of tokens. In particular, they utilize a decoder of a Transformer language model to learn from large amounts of text and image pairs. Parti~\cite{Arxiv2022Jiahui} is a two-stage model similar to DALL-E and CogView, composed of an image tokenizer and an autoregressive model. The first step trains a vision tokenizer VIT-VQGAN~\cite{ICLR2022Jiahui} that transforms an image into a sequence of discrete image tokens. The second step trains an encoder-decoder based Transformer that generates image tokens from text tokens. Parti achieves improved image quality by scaling the encoder-decoder Transformer model up to 20 billion parameters. However, these Auto Regressive Generative models still lack attention to fine-grained level information and require large amounts of data, model size, and training time.

The third approach for text to image synthesis is based on diffusion models, which convert text to image from a learned data distribution by iteratively denoising a learned data distribution. GLIDE~\cite{ICML2022Alexander} was the first work to apply diffusion model with CLIP guidance and classifier-free guidance in text to image synthesis. VQ-Diffusion~\cite{CVPR2022Shuyang} proposed a vector-quantized diffusion model based on VQ-VAE, whose latent space is modeled by a conditional variant of the  Denoising Diffusion Probabilistic Model (DDPM). DALLE-2~\cite{Arxiv2022Aditya} trained a diffusion model on the CLIP image embedding space and a separate decoder to create images based on the CLIP image embeddings. Imagen~\cite{Arxiv2022Chitwan} used a frozen T5-XXL encoder to map text to a sequence of embeddings, an image diffusion model, and two super-resolution image diffusion models. These three image diffusion models are all conditioned on the text embedding sequence and use classifier-free guidance. However, these diffusion models still lack attention to fine-grained level information and require huge resources.

To address the challenge of preserving fine-grained information and minimizing computational costs, we propose that utilizes the Recurrent Affine Transformation (RAT) GAN, which achieved state-of-the-art performance on fine-grained datasets while using acceptable number of parameters. Additionally, we introduce an auxiliary classifier in the discriminator to help RAT GAN synthesize more accurate fine-grained images. Specifically, the classifier classifies both fake and real images and assists the generator in synthesizing fine-grained images. While fine-grained categories may be hard to obtain for images in the wild, they are available in many cases and our approach can leverage this information for improved results. Moreover, semi-supervised and weakly supervised techniques could also help address lack of categories.

Furthermore, we introduce contrastive learning to further improve the fine-grained details of the images synthesized by RAT GAN, particularly on datasets with different subclasses. The contrastive learning method minimizes the similarity of fake/real images from different subclasses and maximizes the similarity of fake/real images from the same subclass. We incorporate the cross-batch memory (XBM)~\cite{CVPR2020Xun}~\cite{CVPR2022Xu} mechanism, which allows the model to collect hard negative pairs across multiple mini-batches and even over the entire dataset, to further improve the performance of the model.

In summary, there are three primary contributions in this paper. First, we introduce an auxiliary classifier in the discriminator, which not only classifies the category of fake/real images but also assists in synthesizing fine-grained images from the generator. Second, we introduce a contrastive learning method with cross-batch memory (XBM) mechanism, which helps the generator to synthesize images with higher similarity within the same subclass and lower similarity among different subclasses. Meanwhile, our method is an efficient approach, as it only introduces small additional expense in the form of two fully connected layers for feature dimension reduction, image classification and feature embedding. Third, our method demonstrates state-of-the-art performance on two common fine-grained image datasets: CUB-200-2011 bird dataset and Oxford-102 flower dataset.

\section{RELATED WORK}

RAT GAN~\cite{Arxiv2022Senmao} was proposed to address text and image isolation during image synthesis. They introduce Recurrent Affine Transformation (RAT) for controlling all fusion blocks consistently. RAT expresses different layers’ outputs with standard context vectors of the same shape to achieve unified control of different layers. The context vectors are then connected using RNN in order to detect long-term dependencies. With skip connections in RNN, RAT blocks are consistent between neighboring blocks and reduce training difficulty. Moreover, they incorporate a spatial attention model in the discriminator to improve semantic consistency between texts and images. With spatial attention, the discriminator can focus on image regions that are related to the corresponding captions. We discovered RAT GAN maintains top performance with acceptable parameters compared to other leading methods. Thus, we adopt the RAT GAN as our backbone model.

The basic GAN framework can be augmented using side information such as class and caption. Instead of feeding side information to the discriminator, one can task the discriminator with reconstructing side information. This is done by modifying the discriminator to contain an auxiliary decoder network that outputs the class label for the training data ~\cite{NIPS2016Tim} or a subset of the latent variables from which are generated ~\cite{NIPS2016Xi}. Forcing a model to perform additional tasks is known to improve performance on the original task. In addition, an auxiliary decoder could leverage pre-trained discriminators for further improving the synthesized images~\cite{NIPS2016Anh}. Motivated by these considerations, ACGAN~\cite{ICML2017Augustus} proposed a class conditional GAN model, but with an auxiliary decoder that is tasked with reconstructing class labels. TAC GAN~\cite{arxiv2017Dash} present a Text Conditioned Auxiliary Classifier Generative Adversarial Network for synthesizing images from their text descriptions. The discriminator of TAC-GAN performs an auxiliary task of classifying the synthesized and the real data into their respective class labels.~\cite{WACV2020Xu} proposed an accelerated WGAN update strategy to speed up the GAN model convergence. ~\cite{CVPR2024Xu} introduced a two-stages training method to fine-grained the image restoration result. Inspired by their work, we introduce an auxiliary classifier in the discriminator of the RAT GAN model. This classifier could not only classify which category the images belong to, but also help generator to synthesize fine-grained level images.

~\cite{BMVC2021Hui} propose a contrastive learning method to improve the quality and enhance the semantic consistency of synthetic images synthesized from texts. In the image-text matching task, they utilize the contrastive loss to minimize the distance of the fake images generated from text descriptions related to the same ground truth image while maximizing those related to different ground truth images. However, they ignored the similarity among fake images of different subclasses and introduced a pretrained image encoder to compute contrastive loss which increased the computation complexity of the model.~\cite{CVPR2019Yin} also propose a contrastive learning method for text to image synthesis. They introduce multiple generators and discriminators and only compute the contrastive loss between image features from the geneartor.~\cite{CVPR2021Zhang} propose a cross-modal contrastive learning for text to image synthesis. They only compute the contrastive loss between a real image and a fake image. In our work, we only add one fully connected layer to extract feature embedding and compute the contrastive loss between fake and real images, between fake and fake images, and between real and real images. The advantage of our approach is that with a small number of parameters, we can compute the contrastive loss between fake/real images in one step, rather than first training an image encoder and then computing contrastive loss as in~\cite{BMVC2021Hui}.

\section{PROPOSED APPROACH}

We adopt the RAT GAN as our base model and enhance it by introducing an auxiliary classifier and a contrastive learning method thus creating a fine-grained (FG) RAT GAN. In the following sections, we provide detailed information on how these modifications work and present the overall algorithms.

\begin{figure*}[t]
\centerline{
\begin{tabular}{cc}
  \resizebox{0.4\textwidth}{!}{{
  \includegraphics{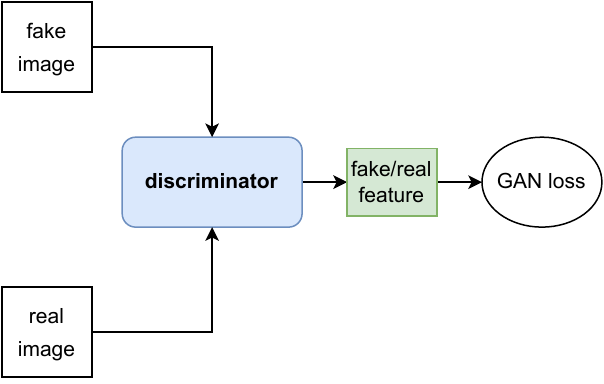}}}
  &
  \resizebox{0.55\textwidth}{!}{{
  \includegraphics{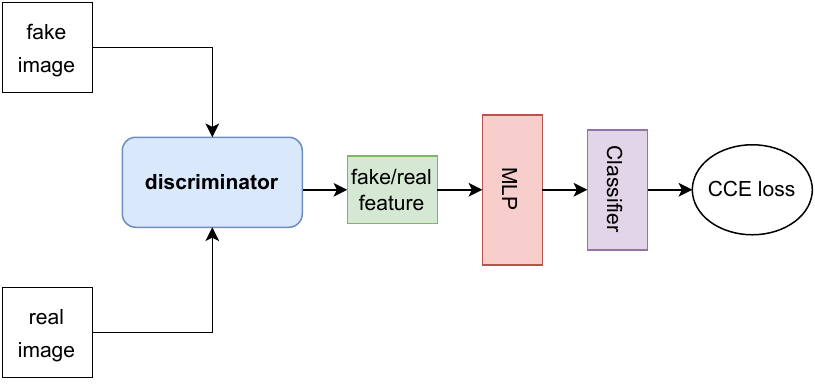}}}
  \\
  (a) original discriminator & (b) discriminator with auxliary classifier
  \\
  \multicolumn{2}{c}{\resizebox{0.95\textwidth}{!}{{
  \includegraphics{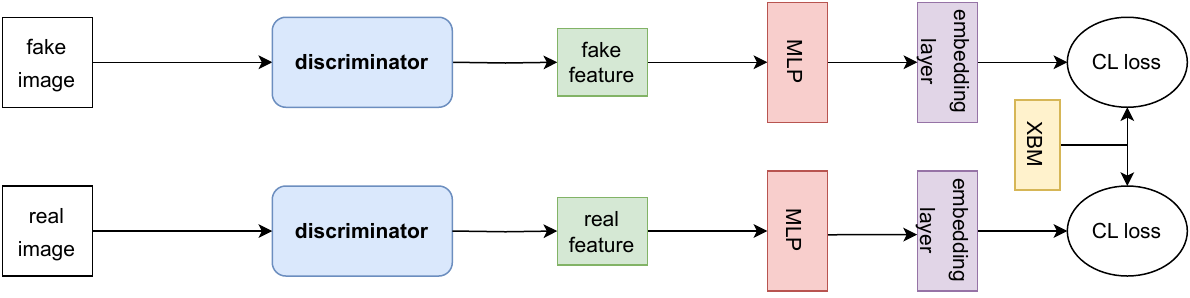}}}}
  \\
%  (a) original discriminator & (b) discriminator with auxliary classifier & (c) discriminator with contrastive learning
\multicolumn{2}{c}{(c) discriminator with contrastive learning}
\end{tabular}}
\caption{The original discrminator in Figure (a) computes GAN loss. The discriminator with auxliary classifier in Figure (b) computes categorical cross entropy loss. The discrminator with contrastive learning in Figure (c) computes contrastive learning loss.
} 
\label{fig: discriminator ablation study}
\end{figure*} 

\subsection{Auxiliary classifier\label{ce loss}}
In the discriminator of the RAT GAN, we add an auxiliary classifier at the end of the network. To do this, we first flatten the output dimension from 8x8x1024 to 64x1024 and add a fully connected layer to reduce the feature dimension from 64x1024 to 256. We then add a Softmax activation function to classify the feature into one of the predefined categories.  The structure of the modified discriminator is shown in Figure~\ref{fig: discriminator ablation study}(b). In comparison to the original RAT GAN discriminator shown in Figure~\ref{fig: discriminator ablation study}(a) which only computes the GAN loss, we also minimize the categorical cross-entropy loss between the classifier output and the ground truth labels of the images during both generator and discriminator updates. These losses are defined as follows:

\begin{equation}
L_d^{ce} = -\sum_{i=1}^{i=N}(y_i \cdot log(\hat{y_i^f})) + \sum_{i=1}^{i=N}(y_i \cdot log(\hat{y_i^r}))
\label{eq: cce d loss}
\end{equation}

\begin{equation}
L_g^{ce} = -\sum_{i=1}^{i=N}(y_i \cdot log(\hat{y_i^f}))
\label{eq: cce g loss}
\end{equation}

where the $y_i$ is the ground truth label of the image, $y_i^{f}$ is the auxiliary output of the fake image, and $y_i^{r}$ is the auxiliary output of the real image. 

$L_d^{ce}$ allows the discriminator to classify the category of images, by computing the sum of the categorical cross-entropy loss between the classifier output of fake images and their ground-truth labels, and the categorical cross-entropy loss between the classifier output of real images and their ground-truth labels. $L_g^{ce}$ helps the generator to synthesize more precise and fine-grained images by incorporating the classifier's output into the loss function.

\subsection{Contrastive learning\label{cl loss}}
In order to improve the quality and semantic consistency of synthetic images generated from text, we introduce a contrastive learning method in our model. To implement this, we add a branch embedding layer and L-2 normalization to the feature embeddings of our images after the fully connected layer for feature dimension reduction. This is illustrated in Figure~\ref{fig: discriminator ablation study}(c).

In addition, we introduce cross-batch memory (XBM) mechanism in our contrastive loss calculation. This creates a memory bank that acts as a queue, where the current mini-batch of real images' feature embeddings are enqueued and the oldest mini-batch of feature embeddings are dequeued. We then minimize the contrastive loss between the fake images' feature embeddings and the entire XBM feature embeddings, as well as the contrastive loss between the real images' feature embeddings and the entire XBM feature embeddings. The contrastive loss is defined as follows:

\begin{equation}
\begin{split}
L_d^{cl}
= &\frac{1}{NM}\sum_{i}^{N}[\sum_{j:y_i=y_j}^{M}(1 - \mbox{cos\_sim}(e_i^r, e_j^x)) + \\
&\sum_{j:y_i \neq y_j}^{M} - \max((\mbox{cos\_sim}(e_i^r, e_j^x) - \alpha), 0)],
\end{split}
\label{eq: cont_d}
\end{equation}

\begin{equation}
\begin{split}
L_g^{cl}
= &\frac{1}{NM}\sum_{i}^{N}[\sum_{j:y_i=y_j}^{M}(1 - \mbox{cos\_sim}(e_i^f, e_j^x)) + \\
&\sum_{j:y_i \neq y_j}^{M} - \max((\mbox{cos\_sim}(e_i^f, e_j^x) - \alpha), 0)],
\end{split}
\label{eq: cont_g}
\end{equation}

where $\mbox{cos\_sim}(e_i^f, e_j^x)$ is the cosine similarity between the feature embedding $e_i^f$ of mini-batched fake images and the feature embedding $e_j^x$ of real image from the cross-batch memory (XBM), $\alpha$ is a margin applied to the cosine similarity of negative pairs to prevent the loss from being dominated by easy negatives, $N$ is the batch size, and $M$ is the size of the XBM. The $L_d^{cl}$ loss function minimizes the similarity between feature embeddings of real images from different subclasses, and maximizes the similarity between feature embeddings of real images from the same subclass, which optimizes the embedding layer in the discriminator. The $L_g^{cl}$ loss function minimizes the similarity between feature embeddings of fake and real images from different subclasses, and maximizes the similarity between feature embeddings of fake and real images from the same subclass, which helps the generator synthesize fine-grained images.

\begin{figure*}[th!]
\centerline{
\begin{tabular}{c}
  \resizebox{0.95\textwidth}{!}{\rotatebox{0}{
  \includegraphics{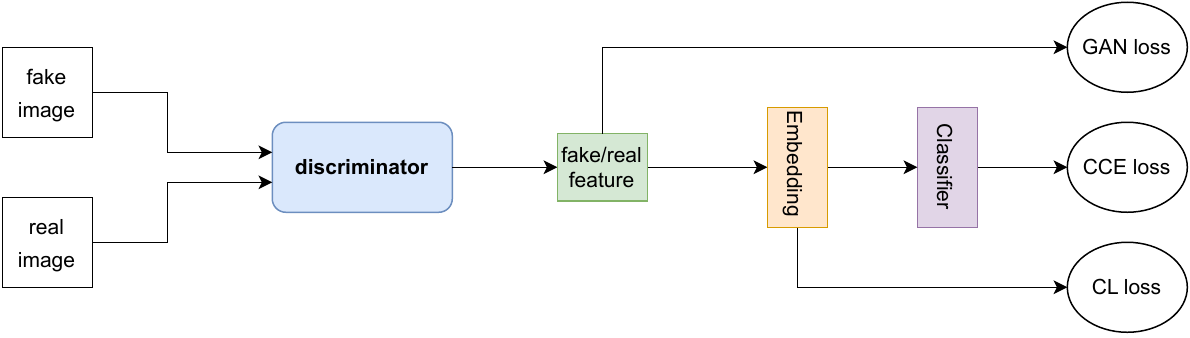}}}
\end{tabular}}
\caption{ The structure of the discriminator with auxiliary classifier and contrastive learning. The original output of the discriminator is still used to compute the GAN loss, and meanwhile followed by one fully connected layer to decrease the feature dimension. Next, the fully connected layer is followed by one embedding layer for contrastive learning. Then, the embedding layer is followed by a classifier for image classification.}
\label{fig: discriminator entire}
\end{figure*} 

\subsection{Training of the network}
In this section, we describe the training process of our proposed FG-RAT GAN with auxiliary classifier and contrastive learning as shown in Figure~\ref{fig: discriminator entire}. The fake image synthesized from generator G and the real image separately pass through discriminator D. The discriminator D then discriminates whether the image is fake/real by minimaxing GAN loss which is defined as follows:

\begin{equation} \label{adv loss of d}
\begin{split}
L_d^{adv}=&\mathbb{E}_{x\backsim p_{data}}[\max{(0, 1-D(i_r, t))}]+\\
&0.5 \times \mathbb{E}_{z\backsim p_{gen}}[\max{(0, 1+D(G(z,t), t))}]+\\
&0.5 \times \mathbb{E}_{z\backsim p_{data}}[\max{(0, 1+D({i_r}^\prime, t))}]
\end{split}
\end{equation}

\begin{equation} \label{adv loss of e}
\begin{split}
L_g^{adv}=\mathbb{E}_{z\backsim p_{gen}}[\min{D(G(z,t), t)}]
\end{split}
\end{equation}

where $G:(Z,T) \rightarrow X$ maps from the latent space Z and caption space T to the input space X, $D:X \rightarrow \mathbb{R}$ maps from the input space to a classification of the example as fake or real, $i_r$ is the real image, and ${i_r}^\prime$ is the mismatched real image. The GAN model will reach a global optimal value when $p_{gen}=p_{data}$, where $p_{gen}$ is the generative data distribution and $p_{data}$ is the real data distribution. 

Subsequently, different from Section~\ref{ce loss} and Section~\ref{cl loss}, in the end of the discriminator, we first add an embedding layer which is used for feature dimension reduction and contrastive learning. We add a classifier after this embedding layer for image classification. We first only compute the categorical cross-entropy loss $L_d^{ce}$ and $L_g^{ce}$ for image classification as mentioned in Section~\ref{ce loss}. This is because the feature drift is relatively large at the early epochs. Training the neural networks with $L_d^{ce}$ and $L_g^{ce}$, allows the embeddings to become more stable. After several training epochs, we add the contrastive loss $L_d^{cl}$ and $L_g^{cl}$ for contrastive learning as mentioned in Section~\ref{cl loss}. We finally compute the total loss for the discriminator D and the generator G as follows:

\begin{equation}
L_d^{total} = L_d^{adv} + L_d^{ce} + L_d^{cl}
\label{eq: d loss}
\end{equation}

\begin{equation}
L_g^{total} = L_g^{adv} + L_g^{ce} + L_g^{cl}
\label{eq: g loss}
\end{equation}

We update the parameters of the discriminator D by minimizing the  $L_d^{total}$ loss and update the parameters of the generator G by minimizing the  $L_g^{total}$ loss.

\section{EXPERIMENTS}

\subsection{Datasets}
To evaluate the performance of fine-grained text to image synthesis, we conduct experiments on two commonly used fine-grained text-image pair datasets: the CUB-200-2011 dataset which contains 11,788 images of 200 different bird species; and the Oxford-102 flower dataset which contains 8,189 images of 102 different flower species. We follow the same split as previous studies~\cite{ICML2016Reed,Arxiv2022Senmao,CVPR2022Shuyang} for both datasets: 150 training classes and 50 testing classes for CUB-200-2011, and 82 training classes and 20 testing classes for Oxford-102. Each image in the datasets is paired with ten text descriptions. The images are resized to 304x304, randomly cropped to 256x256, and then randomly flipped horizontaly. The captions are passed through a text encoder, resulting in an output of size 256.

\subsection{Evaluation metrics}
The Inception Score~\cite{IS} can measure a synthetic image quality by computing the expected Kullback Leibler divergence (KL divergence) between the marginal class distribution and conditional label distribution:
\begin{equation}
IS = exp(\mathbb{E}_xKL(p(y|x)||p(y)))
\end{equation}
where $p(y|x)$ is the conditional label distribution of features extracted from the middle layers of the pretrained Inception-v3 model for generated images, and p(y) is the marginal class distribution. IS gives a score that tells us if each image made by the model is clear and distinct, and if the model can make a wide range of different images. We want models that make a mix of clear images, so a higher IS is better.

The Frechet Inception Distance~\cite{FID} that is given by:
\begin{equation}
d^2(F, G)=|\mu_x-\mu_y|^2+tr|\Sigma_x+\Sigma_y-2(\Sigma_x\Sigma_y)^{1/2}|
\end{equation}
where F, G are two distributions of features extracted from the middle layers of a pretrained Inception-v3 model for generated and real images. The parameters $\mu_x$, $\mu_y$, $\Sigma_x$, $\Sigma_y$, are the mean vectors and covariance matrices of F and G. While IS checks image clarity and variety, FID checks if they look real. We want our model's images to look like real photos, so a lower FID is better. 

The paper~\cite{FIDnote} highlights that the Inception Score (IS) is sensitive to model overfitting and dependent on the dataset used for the Inception network, often leading to misleading evaluations for models not trained on ImageNet. In contrast, the Frechet Inception Distance (FID) compares the statistical distributions of real and generated images using the Frechet distance, assessing how closely generated images mimic real images in content and style. This makes FID a more reliable and comprehensive metric, as it directly evaluates the realism and diversity of generated images, unlike IS which does not compare with the distribution of real images.

\begin{figure}[htb!]
\centering
\begin{tabular}{cccccc}
\hline
 Class & Target & LAFITE & VQ-Diffusion & RAT GAN & FG-RAT GAN
  \\ \hline
    \shortstack{Class 001 \\
  Black Footed  \\
  Albatross  \\
 $0001\_796111.png$
    \\ \qquad \\ \qquad \\ \qquad}
  &
  \includegraphics[width=0.15\linewidth]{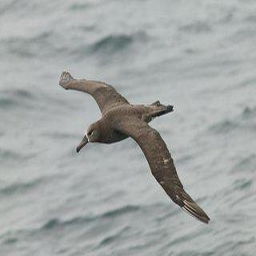}
  &
  \includegraphics[width=0.15\linewidth]{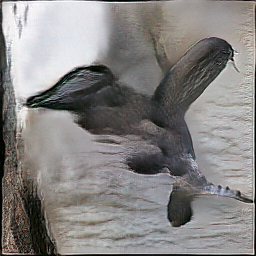}
  &
  \includegraphics[width=0.15\linewidth]{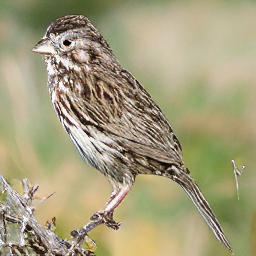}
  &
  \includegraphics[width=0.15\linewidth]{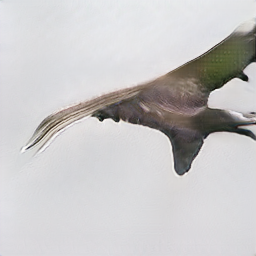}
  &
  \includegraphics[width=0.15\linewidth]{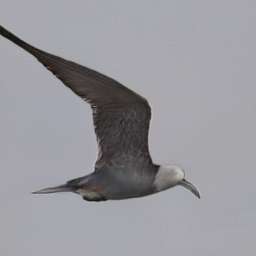}
  \\
\multicolumn{6}{c}{
  \begin{tabular}{@{}c@{}}
    caption: the entire body is dark brown with a white band encircling\\
     where the bill meets the head.
  \end{tabular}
}
  \\ \hline 
     \shortstack{Class 001 \\
  Black Footed  \\
  Albatross  \\
 $0002\_55.png$
    \\ \qquad \\ \qquad \\ \qquad}
   &
  \includegraphics[width=0.15\linewidth]{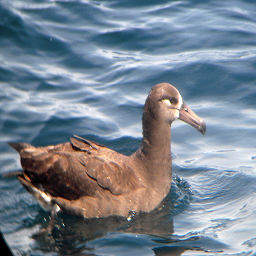}
  &
  \includegraphics[width=0.15\linewidth]{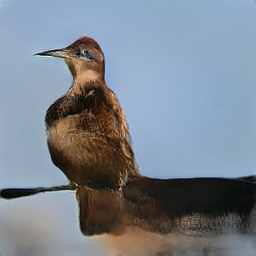}
  &
  \includegraphics[width=0.15\linewidth]{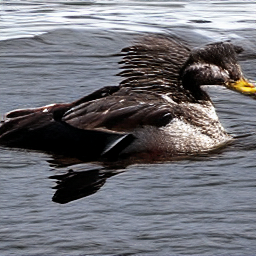}
    &
  \includegraphics[width=0.15\linewidth]{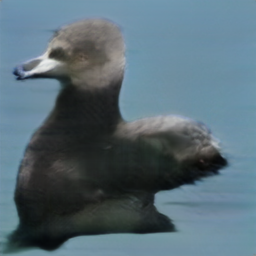}
    &
  \includegraphics[width=0.15\linewidth]{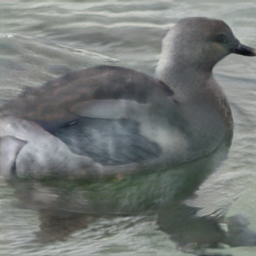}
  \\
   \multicolumn{6}{c}{caption: this bird has wings that are brown and has a big bill.}
   \\ \hline 
    \shortstack{Class 001 \\
  Black Footed  \\
  Albatross  \\
 $0005\_796090.png$
    \\ \qquad \\ \qquad \\ \qquad}
   &
  \includegraphics[width=0.15\linewidth]{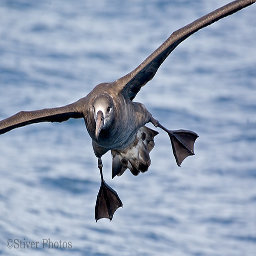}
  &   
  \includegraphics[width=0.15\linewidth]{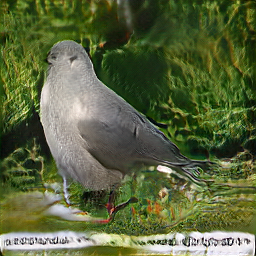}
  &
  \includegraphics[width=0.15\linewidth]{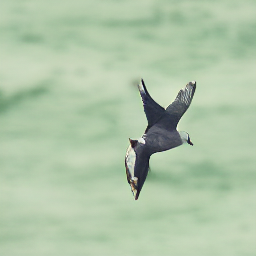}
  &
  \includegraphics[width=0.15\linewidth]{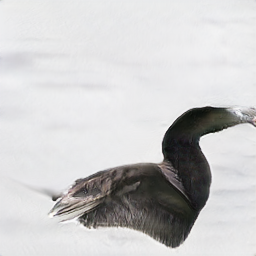}
  &
  \includegraphics[width=0.15\linewidth]{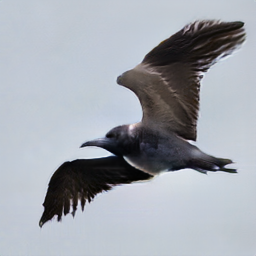}
    \\
 %\multicolumn{6}{c}{
  %\begin{tabular}{@{}c@{}}
   % caption: this bird has large feet and a broad wingspan \\
   %with all grey coloration.
  %\end{tabular}
%}
   \multicolumn{6}{c}{caption: this bird has large feet and a broad wingspan with all grey coloration.}
    \\ \hline
  \shortstack{Class 014 \\
  Indigo  \\
  Bunting  \\
 $0001\_12469.png$
    \\ \qquad \\ \qquad \\ \qquad}
    &
  \includegraphics[width=0.15\linewidth]{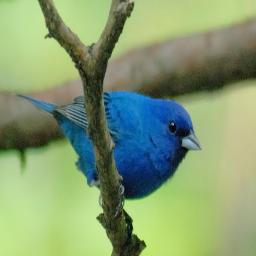}
  &     
  \includegraphics[width=0.15\linewidth]{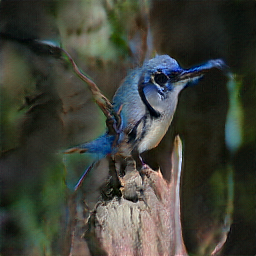}
    &
  \includegraphics[width=0.15\linewidth]{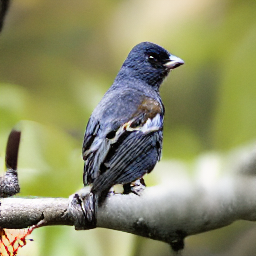}
    &
  \includegraphics[width=0.15\linewidth]{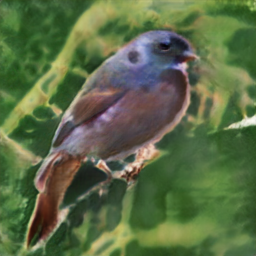}
    &
  \includegraphics[width=0.15\linewidth]{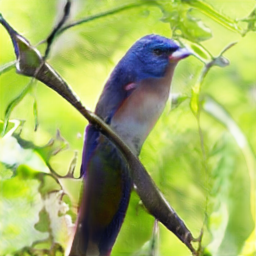}
  \\
%      \multicolumn{6}{c}{
 % \begin{tabular}{@{}c@{}}
  %  caption: this bird has a short, pointed blue beak, \\
 %   it also has a blue tarsus and blue feet.
 % \end{tabular}
%}
     \multicolumn{6}{c}{caption: this bird has a short, pointed blue beak, it also has a blue tarsus and blue feet.}
    \\ \hline 
  \shortstack{Class 014 \\
  Indigo  \\
  Bunting  \\
 $0047\_12966.png$
    \\ \qquad \\ \qquad \\ \qquad}
  &
  \includegraphics[width=0.15\linewidth]{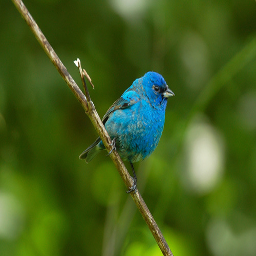}
  &
  \includegraphics[width=0.15\linewidth]{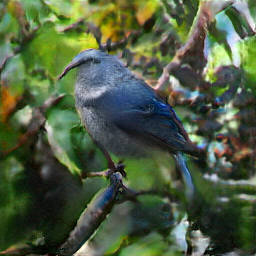}
    &
  \includegraphics[width=0.15\linewidth]{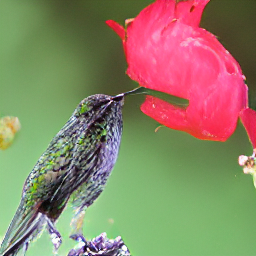}
    &
  \includegraphics[width=0.15\linewidth]{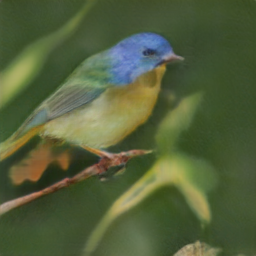}
    &
  \includegraphics[width=0.15\linewidth]{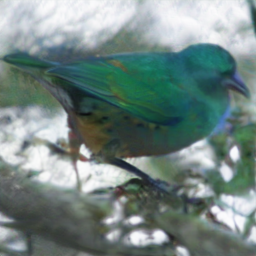}
    \\
    \multicolumn{6}{c}{
  \begin{tabular}{@{}c@{}}
    caption: a small colorful bird with teal feathers covering its body, \\
    with green speckles on its vent and abdomen.
  \end{tabular}
}
    % \multicolumn{6}{c}{caption: a small colorful bird with teal feathers covering its body, with green speckles on its vent and abdomen.}
    \\ \hline
   \shortstack{Class 014 \\
  Indigo  \\
  Bunting  \\
 $0059\_11596.png$
    \\ \qquad \\ \qquad \\ \qquad}
   &
  \includegraphics[width=0.15\linewidth]{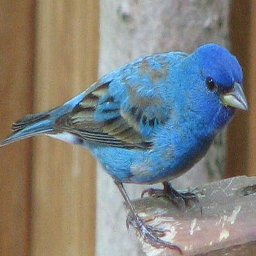}
  &   
  \includegraphics[width=0.15\linewidth]{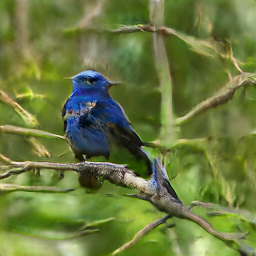}
  &
  \includegraphics[width=0.15\linewidth]{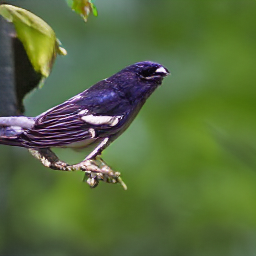}
    &
  \includegraphics[width=0.15\linewidth]{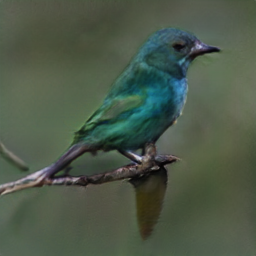}
    &
  \includegraphics[width=0.15\linewidth]{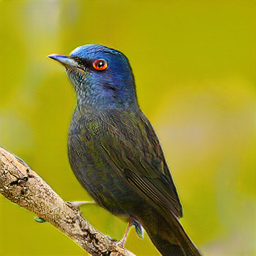}
   \\
     \multicolumn{6}{c}{caption: a small purple bird, with black primaries, and a thick bill.}
    \\ \hline 
\end{tabular}
\caption{Examples of generated images using RAT GAN and the proposed FG-RAT GAN on the CUB bird dataset. Each row represents a different sample (image size = 256x256) and with the corresponding caption below.The first column is image class and name. The second column is the corresponding target image. The rest of other columns are the generated images from LAFITE, VQ-Diffusion, RAT GAN, and our FG-RAT GAN.  As we can see, our FG-RAT GAN can generate more realistic images where each image is similar to other images within the same class.}
\label{fig: bird}
\end{figure} 

\begin{figure*}[htb!]
\center{
\begin{tabular}{ccccc}
\hline
 Class & Caption & Target & RAT GAN & FG-RAT GAN
  \\ \hline
  \shortstack{Class 032 \\
 $image\_05587.png$
 \\ \qquad \\ \qquad \\ \qquad
 \\ \qquad \\ \qquad \\ \qquad}
  &
  \shortstack{the petals of flowers\\
  are various shades\\
  of pink and have\\
  five individual petals.
  \\ \qquad \\ \qquad \\ \qquad}
  &
  \resizebox{0.15\textwidth}{!}{\rotatebox{0}{
  \includegraphics{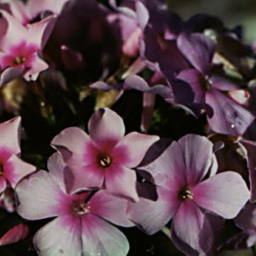}}}
  &
  \resizebox{0.15\textwidth}{!}{\rotatebox{0}{
  \includegraphics{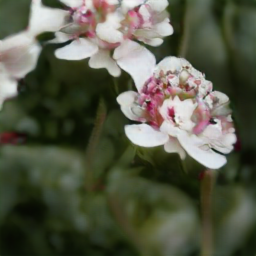}}}
  &
  \resizebox{0.15\textwidth}{!}{\rotatebox{0}{
  \includegraphics{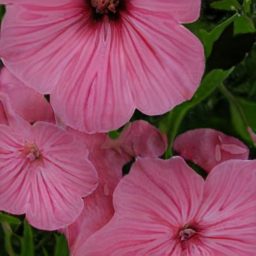}}}\\
  \\ \hline 
  \shortstack{Class 032 \\
 $image\_05602.png$
 \\ \qquad \\ \qquad \\ \qquad
 \\ \qquad \\ \qquad \\ \qquad}
    &
  \shortstack{ a large group of \\
  light pink flowers \\
  with dark pink centers.
  \\ \qquad \\ \qquad \\ \qquad}
  &
  \resizebox{0.15\textwidth}{!}{\rotatebox{0}{
  \includegraphics{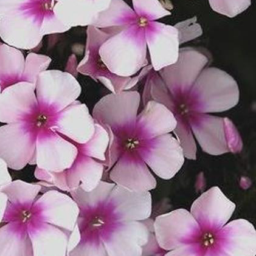}}}
    &
  \resizebox{0.15\textwidth}{!}{\rotatebox{0}{
  \includegraphics{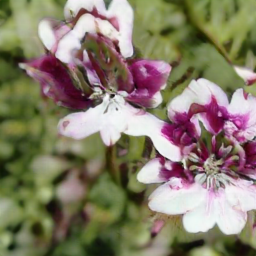}}}
    &
  \resizebox{0.15\textwidth}{!}{\rotatebox{0}{
  \includegraphics{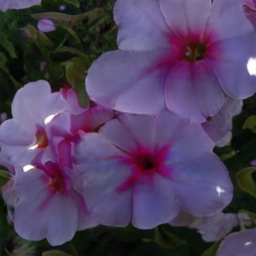}}}
   \\ \hline 
  \shortstack{Class 032 \\
 $image\_05604.png$
 \\ \qquad \\ \qquad \\ \qquad
 \\ \qquad \\ \qquad \\ \qquad}
    &
  \shortstack{these flowers are \\
  mostly pink but \\
  some of them have \\
  white parts located \\
  closer to their stamens.
  \\ \qquad \\ \qquad \\ \qquad}
  &
  \resizebox{0.15\textwidth}{!}{\rotatebox{0}{
  \includegraphics{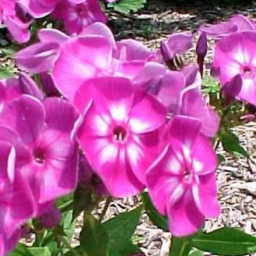}}}
    &
  \resizebox{0.15\textwidth}{!}{\rotatebox{0}{
  \includegraphics{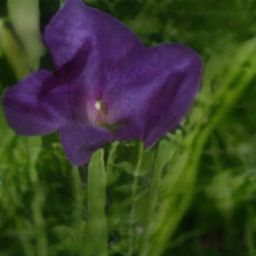}}}
    &
  \resizebox{0.15\textwidth}{!}{\rotatebox{0}{
  \includegraphics{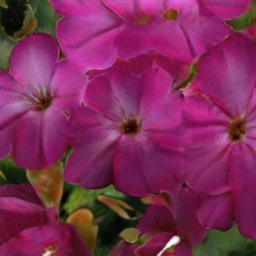}}}
    \\ \hline 
  \shortstack{Class 049 \\
 $image\_06209.png$
    \\ \qquad \\ \qquad \\ \qquad}
    &
  \shortstack{this flower has \\
  thin white petals \\
  as its main feature.
   \\ \qquad \\ \qquad \\ \qquad}
  &
  \resizebox{0.15\textwidth}{!}{\rotatebox{0}{
  \includegraphics{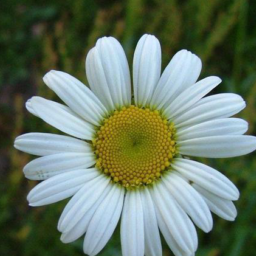}}}
    &
  \resizebox{0.15\textwidth}{!}{\rotatebox{0}{
  \includegraphics{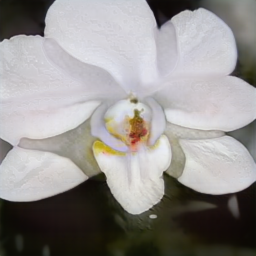}}}
    &
  \resizebox{0.15\textwidth}{!}{\rotatebox{0}{
  \includegraphics{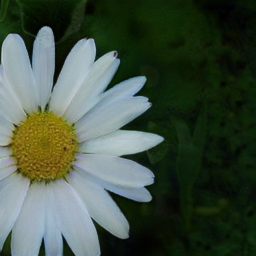}}}
    \\ \hline 
  \shortstack{Class 049 \\
 $image\_06216.png$
    \\ \qquad \\ \qquad \\ \qquad}
    &
  \shortstack{the petals on \\
  this flower are white \\
  with yellow stamen.
      \\ \qquad \\ \qquad \\ \qquad}
  &
  \resizebox{0.15\textwidth}{!}{\rotatebox{0}{
  \includegraphics{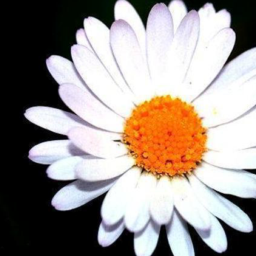}}}
    &
  \resizebox{0.15\textwidth}{!}{\rotatebox{0}{
  \includegraphics{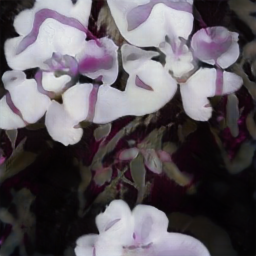}}}
    &
  \resizebox{0.15\textwidth}{!}{\rotatebox{0}{
  \includegraphics{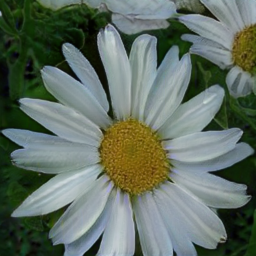}}}
    \\ \hline
  \shortstack{Class 049 \\
 $image\_06224.png$
    \\ \qquad \\ \qquad \\ \qquad}
    &
  \shortstack{the flower has petals \\
  of a white color with\\
  a many yellow stamen.
      \\ \qquad \\ \qquad \\ \qquad}
  &
  \resizebox{0.15\textwidth}{!}{\rotatebox{0}{
  \includegraphics{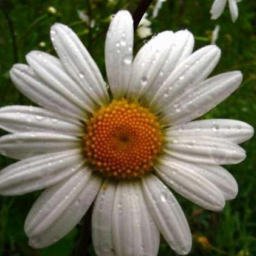}}}
    &
  \resizebox{0.15\textwidth}{!}{\rotatebox{0}{
  \includegraphics{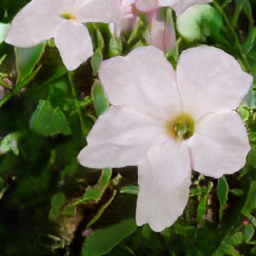}}}
    &
  \resizebox{0.15\textwidth}{!}{\rotatebox{0}{
  \includegraphics{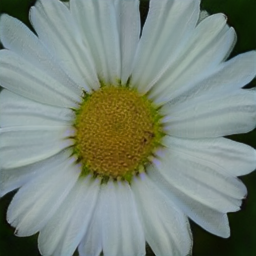}}}
    \\ \hline 
\end{tabular}}
\caption{Examples of generated images using RAT GAN and the proposed FG-RAT GAN with classifier and contrastive learning trained on the Oxford flower dataset. Each row represents a different sample (image size=256x256). The first column is the sample detail including class and specific image name. The second column is the caption. The third column is the corresponding target image. The fourth column is the image generated by RAT GAN. The fifth column is the image generated by our proposed FG-RAT GAN. As we can see, our proposed FG-RAT GAN can generate more realistic images where each image is similar to other images within the same class.}
\label{fig: flower}
\end{figure*} 

\subsection{Implementation details}
In our implementation, we adopt the RAT GAN architecture as the backbone for our model. We use a pretrained bidirectional LSTM network to convert text descriptions into sentence-level feature vectors of size 256. These feature vectors are combined with Gaussian noise as input for the generator. The generator comprises of six up-sampling blocks, each followed by a Recurrent Affine Transformation (RAT) block to control image content. The discriminator includes six down-sampling blocks, whose output size is 8x8x1024. We then add a fully connected layer to decrease the output size to 256 for contrastive learning, followed by a fully connected layer for image classification with an output size of 200 for the CUB-200-2011 dataset and 102 for the Oxford-102 dataset. We use the Adam optimizer to train the generator with an initial learning rate of $1e-4$ and the discriminator with an initial learning rate of $4e-4$. We use cosine learning rate decay to decrease the learning rate to $1e-6$ and train with 600 epochs.

\subsection{Qualitative evaluation}
Figure~\ref{fig: bird} shows synthesized images generated by LAFITE, VQ-Diffusion, RAT GAN and our FG-RAT GAN on the CUB-200-2011 bird dataset.  As we can see, in the 1st row the proposed FG-RAT GAN generates a bird with dark brown body and white band encircling near the bill as specified in the caption, in the 3rd row it generates a bird with all gray body as specified in the caption, and both examples are similar to each other given that they belong to the same class. Figure~\ref{fig: flower} only shows synthesized images generated by RAT GAN and our proposed FG-RAT GAN on the Oxford-102 flower dataset since LAFITE did not train or test  on this dataset and VQ-Diffusion did not post their pretrianed model on this dataset. As we can see, the 5th row generates a flower with white petals and yellow stamen as in the description, the 6th row generates a flower with white petals and yellow stamen as in the description, and both samples are similar to each other given they belong to the same class. There are six samples which belong to two different classes in each dataset. As we can see, our proposed FG-RAT GAN can generate fine-grained images which highly correspond to the given captions. Additionally, each synthesized image is more similar to other synthesized images in the same class. Thus, we demonstrate that our FG-RAT GAN can reach better visualized results compared with the orginal RAT GAN. In addition, we  show some visualized results compared with DALLE-2 and Stable Diffusion on these datasets in the Supplementary materials.\footnotemark[\value{footnote}]

\begin{table*}[htb!]
\begin{center}
{
\begin{tabular}{|c|c||c|c||c|c|}
\hline
  \multicolumn{2}{|c||}{} & \multicolumn{2}{c||}{CUB bird dataset} & \multicolumn{2}{c|}{Oxford flower dataset}\\
\hline
Model & NP & IS$\uparrow$ & FID$\downarrow$ & IS$\uparrow$ & FID$\downarrow$ \\
\hline \hline
LAFITE & 75M+151M & \textcolor{red}{$5.97$} & $10.48$ & $-$ & $-$\\
\hline
VQ-Diffusion & 370M & $-$ & $10.32$ & $-$ & $14.1$\\
\hline
RAT GAN & 38M+113M & $4.83$ & $12.12$ & \textcolor{red}{$3.62$} & $12.90$\\
\hline
FG-RAT GAN \textbf{(our)}& 38M+130M & $4.99$ & \textcolor{red}{$8.66$} & $3.45$ & \textcolor{red}{$9.14$}\\
\hline
\end{tabular}
}
\end{center}
\caption{Comparison of previous state-of-the-art methods: LAFITE, VQ-Diffusion, RAT GAN and our proposed FG-RAT GAN on the CUB-200-2011 bird and Oxford-102 flower dataset for text to image synthesis. Each row presents a different model. The first column is the name of each model. The second column is the number of parameters of each model. The third and forth columns show the IS and FID results for the bird dataset. The fifth and sixth columns show the IS and FID results for the flower dataset.$''-''$ means the author did not provide results. As can be observed, in both datasets, our proposed FG-RAT GAN reaches the lowest FID scores.}
\label{tab: bird and flower}
\end{table*}

\subsection{Quantitative evaluation}
We compare the state-of-the-art text to image synthesis methods LAFITE, VQ-Diffusion, RAT GAN, and our FG-RAT GAN. We evaluate the CUB-200-2011 bird dataset and the Oxford-102 flower dataset with Inception Score (IS) and Frenchet Inception Distance (FID) which are commonly used text to image synthesis performance evaluation metrics. Due to suboptimalities of the Inception Score itself and problems with the popular usage of the Inception Score, we care more about FID than IS. We show the evaluation results in Table~\ref{tab: bird and flower}. As can be observed, on the CUB-200-2011 bird dataset, our method reaches the lowest FID scores. On the Oxford flower dataset, RAT GAN reaches the highest IS score and our proposed method reaches the lowest FID score. In addition, our proposed method only add 17M parameters to the discriminator of RAT GAN and has 168M parameters while LAFITE has 226M parameters and VQ-Diffusion has 370M parameters. Even though we use labels during the training, label information is not an unfair advantage but a distinct characteristic of our model. The goal is to advance the field, rather than to compete under identical conditions. Thus, we demonstrate our FG-RAT GAN reaches better performance while only adding a relatively small number parameters to the baseline model.

\begin{table*}[htb!]
\begin{center}
{
\begin{tabular}{|c||c|c||c|c|}
\hline
  \multicolumn{1}{|c||}{} & \multicolumn{2}{c||}{CUB bird dataset} & \multicolumn{2}{c|}{Oxford flower dataset}\\
\hline
 Model & IS$\uparrow$ & FID$\downarrow$ & IS$\uparrow$ & FID$\downarrow$\\
\hline \hline
RAT GAN & $4.83$ & $12.12$ & $3.62$ & $12.90$\\
\hline
RAT GAN + classifier \textbf{(our)}& \textcolor{red}{$5.08$} & $9.90$ & $3.45$ & $9.55$\\
\hline
RAT GAN + contrtastive learning \textbf{(our)}& $4.84$ & $9.10$ & \textcolor{red}{$3.66$} & $10.63$\\
\hline
FG-RAT GAN \textbf{(our)} & $4.99$ & \textcolor{red}{$8.66$} & $3.45$ & \textcolor{red}{$9.14$}\\
\hline
\end{tabular}
}
\end{center}
\caption{Comparison of RAT GAN, proposed FG-RAT GAN with auxiliary classifier, proposed FG-RAT GAN with contrastive learning, and proposed FG-RAT GAN with combination of auxiliary classifier and contrastive learning on the CUB-200-2011 bird and Oxford-102 flower dataset. Each row presents a different model. The first column is the name of each model. The second and third columns show the IS and FID scores for the CUB bird dataset. The fourth and fifth columns show the IS and FID scores for the Oxford flower dataset. As can be observed, in CUB bird dataset, the proposed FG-RAT GAN with classifier reaches the highest IS score and the proposed FG-RAT GAN with classifier and contrastive learning reaches the lowest FID score. In the Oxford flower dataset, the proposed FG-RAT GAN with contrastive learning reaches the highest IS and the proposed FG-RAT GAN with classifier and contrastive learning reaches the lowest FID.}
\label{tab: ablation bird and flower}
\end{table*}

\subsection{Ablation study}
We investigate the effects of different strategies we added to the RAT GAN model for text to image synthesis to demonstrate their significance on both the CUB-200-2011 bird and Oxford-102 flower datasets. We train three different models: A proposed FG-RAT GAN with auxiliary classifier, a proposed FG-RAT GAN with contrastive learning, and a proposed FG-RAT GAN with combination of auxiliary classifier and contrastive learning. The results are summarized in the Table~\ref{tab: ablation bird and flower}. As can be observed, the proposed FG-RAT GAN with auxiliary classifier reaches the highest IS score, whereas the proposed FG-RAT GAN with a combination of auxiliary classifier and contrastive learning reaches the lowest FID score on the CUB-200-2011 bird dataset. The proposed FG-RAT GAN with contrastive learning reaches the highest IS score and the proposed FG-RAT GAN with combination of auxiliary classifier and contrastive learning reaches the lowest FID score on the Oxford-102 flower dataset. In summary, the ablation study demonstrates that our FG-RAT GAN reaches better performance than the RAT GAN model.

\section{Conclusion}
In this paper, we present a novel approach for generating fine-grained images from text descriptions, by incorporating an auxiliary classifier and contrastive learning into the RAT GAN architecture. Our proposed FG-RAT GAN approach improves the quality and semantic consistency of synthetic images by leveraging the auxiliary classifier to classify images into different categories, and using contrastive learning to generate images with higher similarity within the same class and lower similarity among different classes. Additionally, our method is computationally efficient, as it adds two fully connected layers to the original RAT GAN model only during training stage. We demonstrate that our method reaches state-of-the-art performance on two commonly used fine-grained image datasets. While FG-RAT GAN demonstrates strong performance, it does depend on the availability of fine-grained labels, which could limit its applicability in real-world scenarios where labels are less accurate or unavailable. In future work, we aim to reduce this dependency and explore the method's adaptability in more diverse and less structured environments. Additionally, we will conduct further evaluations on broader text-to-image synthesis benchmarks and more varied datasets are necessary to confirm the generalizability of our approach.


%%%%%%%%% BODY TEXT
\section{Appendix}

\subsection{Comparision results}
We compare with the DALLE-2 and Stable Diffusion which are the most popular models for text to image synthesis task. Since neither DALLE-2 nor Stable Diffusion did not train on the CUB-200-2011 bird dataset and Oxford-102 flower dataset, we only show the visualized results in Figure~\ref{fig: bird} and in Figure~\ref{fig: flower}.

Figure~\ref{fig: bird} and  Figure~\ref{fig: flower} show synthesized images generated by DALLE-2, Stable Diffusion, and our proposed FG-RAT GAN on the bird and flower dataset. There are six samples which belong to two different classes in each dataset. As we can see, our proposed FG-RAT GAN can generate fine-grained images which highly correspond to the given captions. Additionally, each synthesized image is more similar to other synthesized images in the same class. Thus, we demonstrate that our proposed FG-RAT GAN can reach better visualized results compared with DALLE-2 and Stable Diffusion.

\begin{figure}[htbp]
\centering
\begin{tabular}{ccccc}
\hline
Class & Target & DALLE-2 & Stable Diffusion &  FG-RAT GAN \\ \hline
\shortstack{Class 001 \\ Black Footed \\ Albatross \\ $0001\_796111.png$ \\ \qquad \\ \qquad \\ \qquad} &
\includegraphics[width=0.18\textwidth]{pic/bird/Black_Footed_Albatross_0001_796111.png} &
\includegraphics[width=0.18\textwidth]{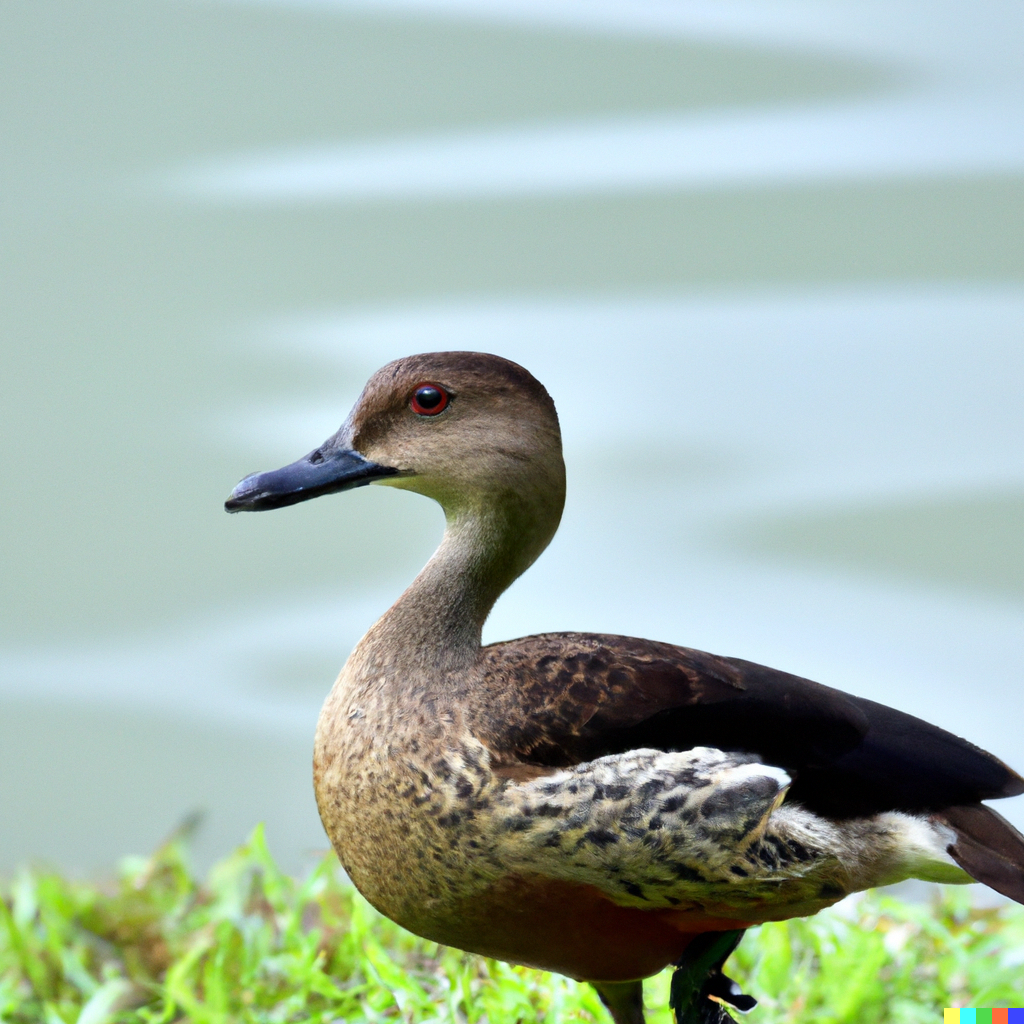} &
\includegraphics[width=0.18\textwidth]{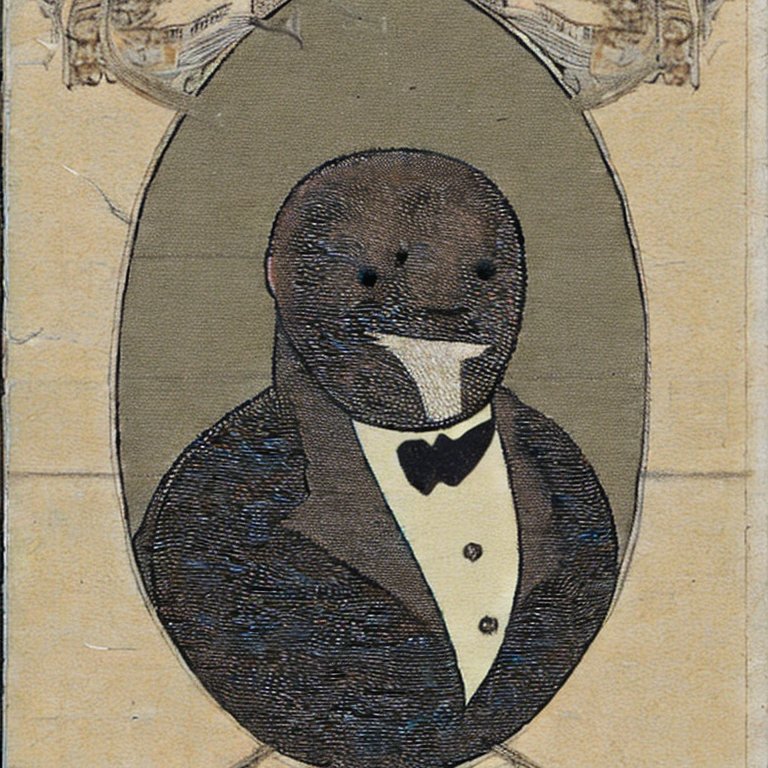} &
\includegraphics[width=0.18\textwidth]{pic/bird/Black_Footed_Albatross_0001_796111_f2.png} \\
\multicolumn{5}{p{1.0\textwidth}}{\textbf{caption:} the entire body is dark brown with a white band encircling where the bill meets the head.} \\ \hline

\shortstack{Class 001 \\ Black Footed \\ Albatross \\ $0002\_55.png$ \\ \qquad \\ \qquad \\ \qquad} &
\includegraphics[width=0.18\textwidth]{pic/bird/Black_Footed_Albatross_0002_55.png} &
\includegraphics[width=0.18\textwidth]{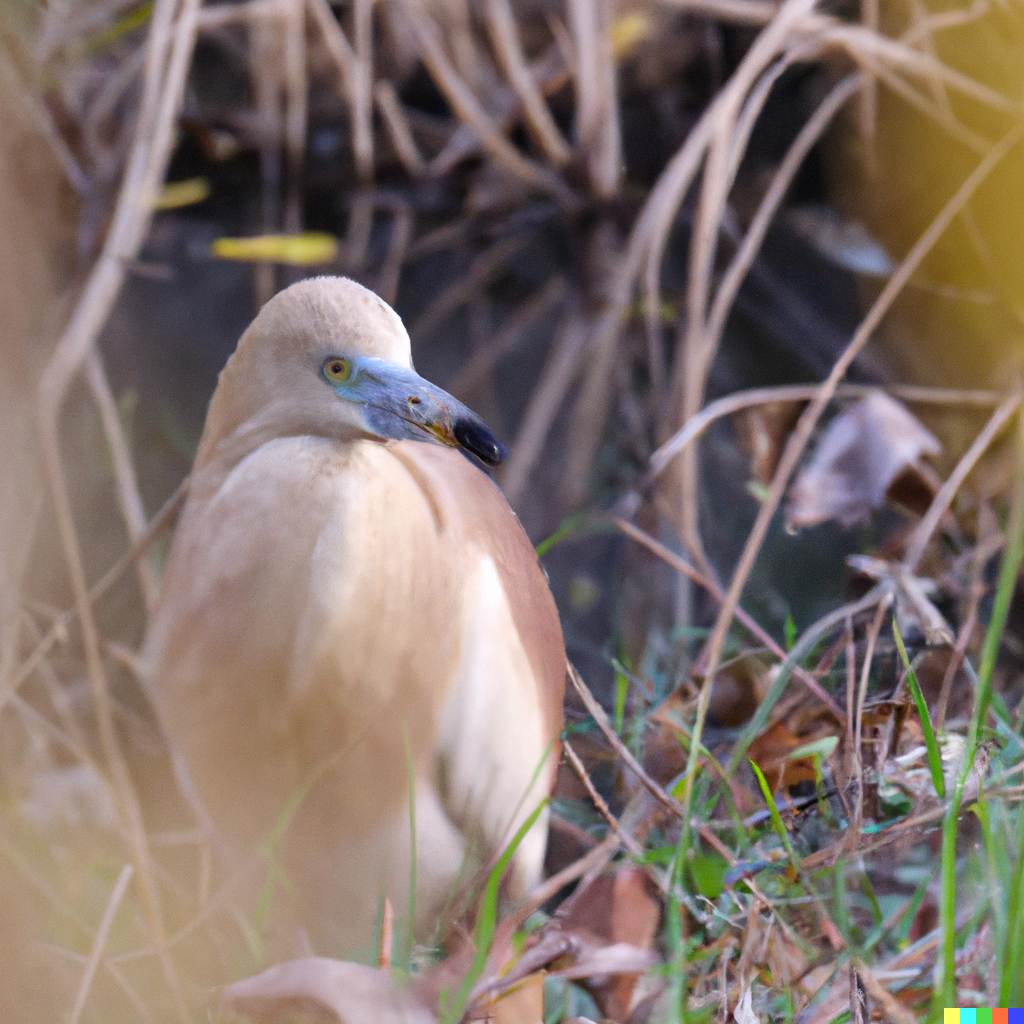} &
\includegraphics[width=0.18\textwidth]{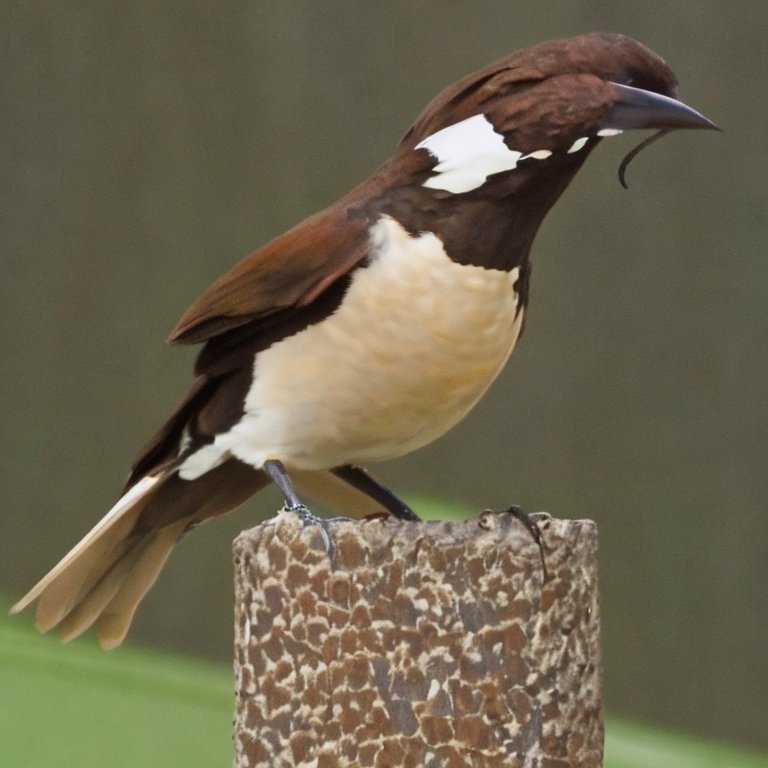} &
\includegraphics[width=0.18\textwidth]{pic/bird/Black_Footed_Albatross_0002_55_f2.png} \\
\multicolumn{5}{p{1.0\textwidth}}{\textbf{caption:} this bird has wings that are brown and has a big bill.} \\ \hline

\shortstack{Class 001 \\ Black Footed \\ Albatross \\ $0005\_796090.png$ \\ \qquad \\ \qquad \\ \qquad} &
\includegraphics[width=0.18\textwidth]{pic/bird/Black_Footed_Albatross_0005_796090.png} &
\includegraphics[width=0.18\textwidth]{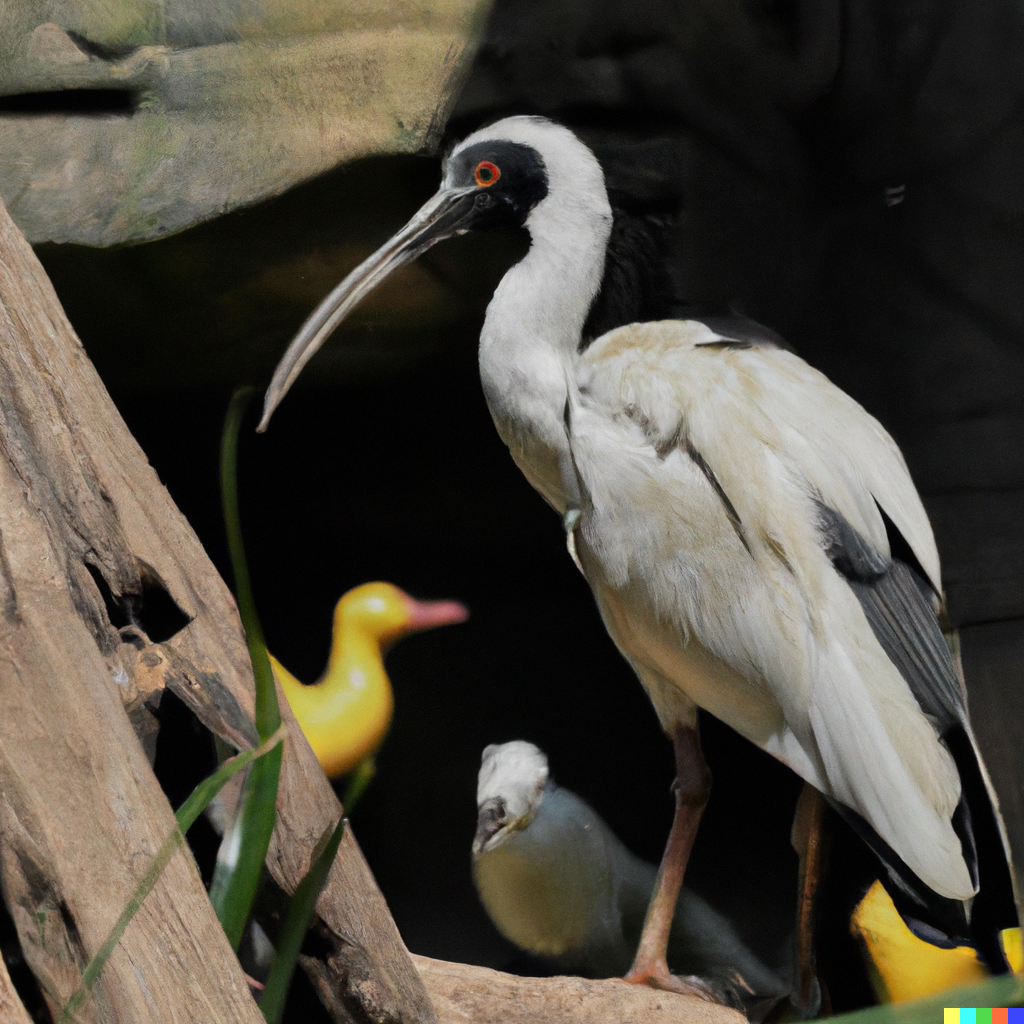} &
\includegraphics[width=0.18\textwidth]{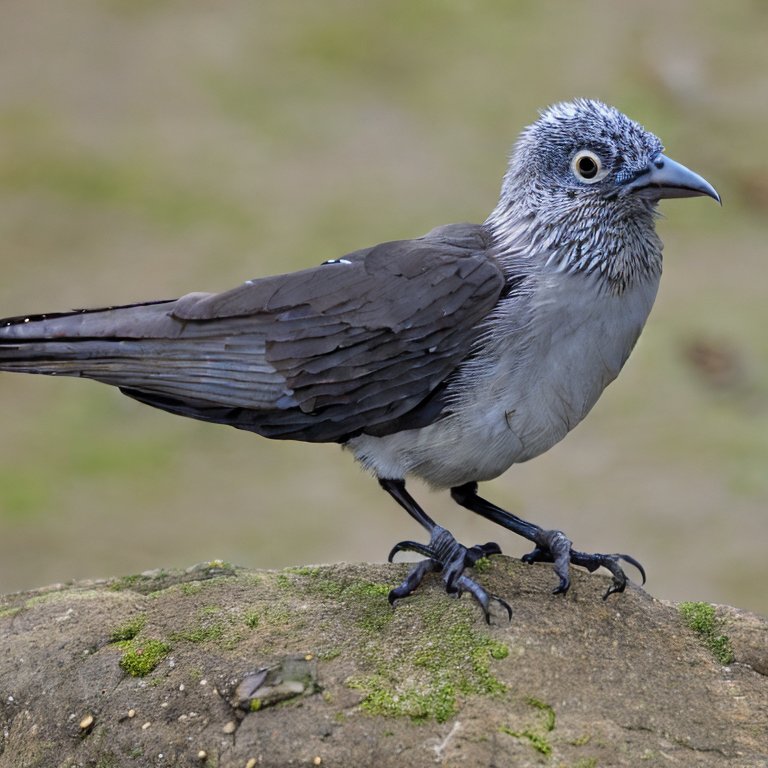} &
\includegraphics[width=0.18\textwidth]{pic/bird/Black_Footed_Albatross_0005_796090_f2.png} \\
\multicolumn{5}{p{1.0\textwidth}}{\textbf{caption:} this bird has large feet and a broad wingspan with all grey coloration.} \\ \hline

\shortstack{Class 014 \\ Indigo \\ Bunting \\ $0001\_12469.png$ \\ \qquad \\ \qquad \\ \qquad} &
\includegraphics[width=0.18\textwidth]{pic/bird/Indigo_Bunting_0001_12469.png} &
\includegraphics[width=0.18\textwidth]{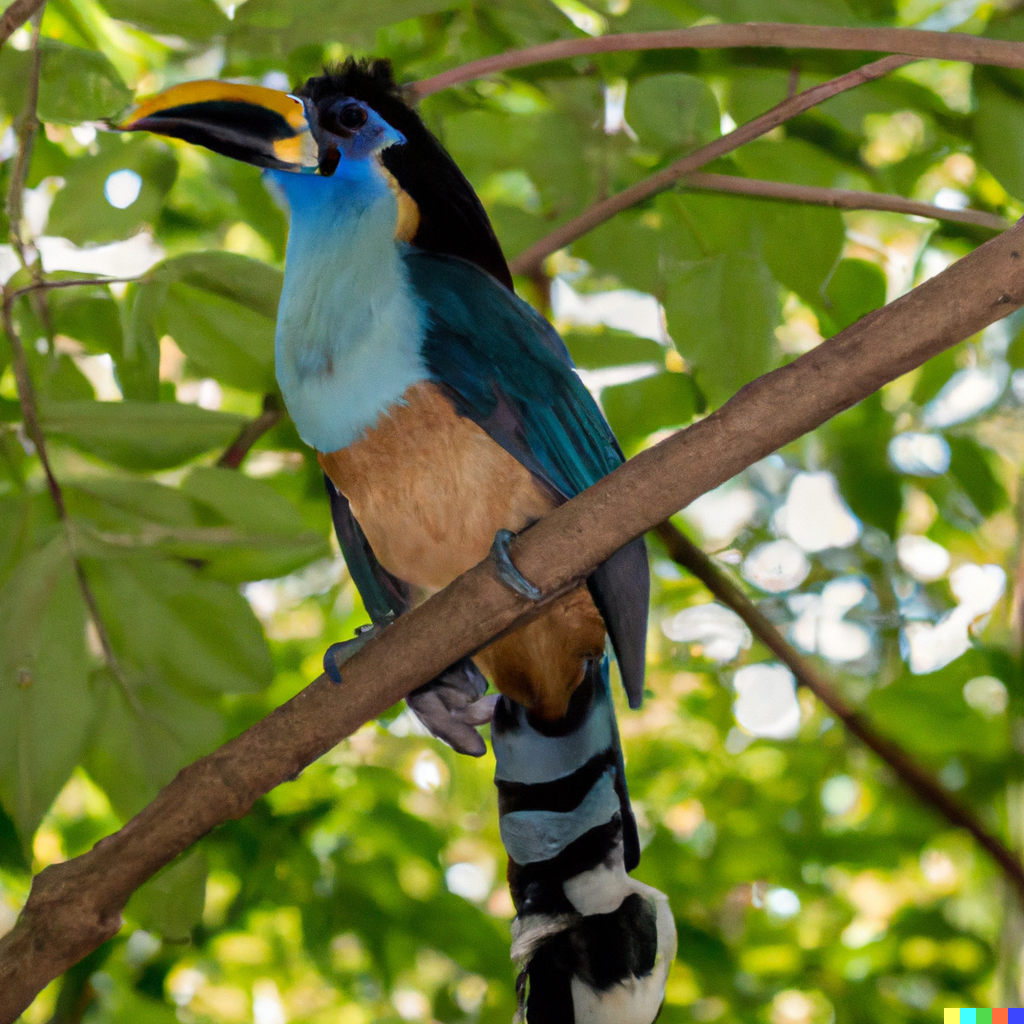} &
\includegraphics[width=0.18\textwidth]{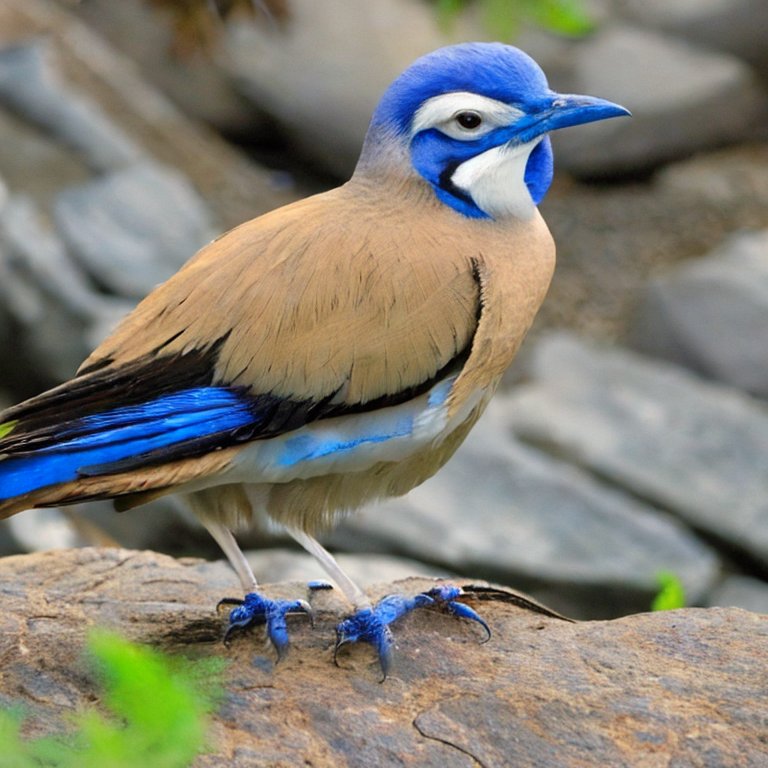} &
\includegraphics[width=0.18\textwidth]{pic/bird/Indigo_Bunting_0001_12469_f2.png} \\
\multicolumn{5}{p{1.0\textwidth}}{\textbf{caption:} this bird has a short, pointed blue beak, it also has a blue tarsus and blue feet.} \\ \hline

\shortstack{Class 014 \\ Indigo \\ Bunting \\ $0047\_12966.png$ \\ \qquad \\ \qquad \\ \qquad} &
\includegraphics[width=0.18\textwidth]{pic/bird/Indigo_Bunting_0047_12966.png} &
\includegraphics[width=0.18\textwidth]{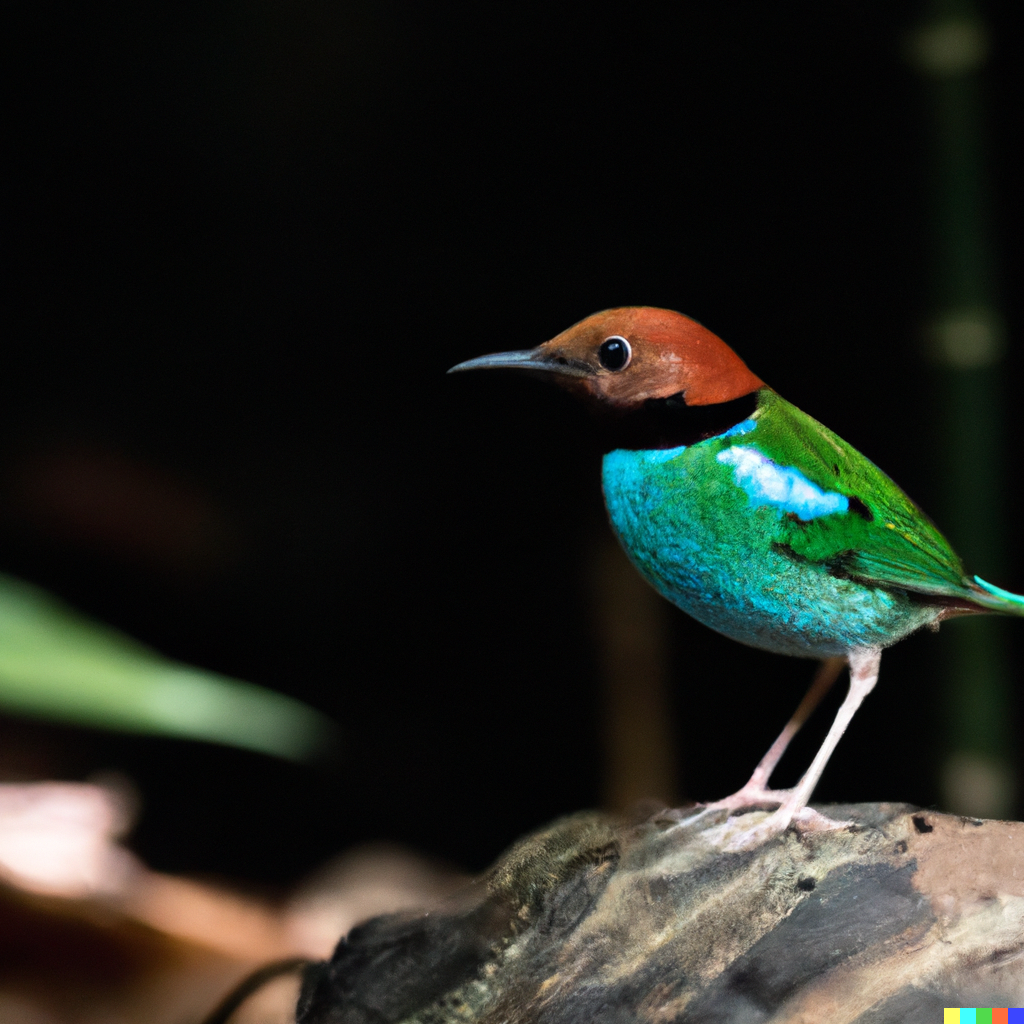} &
\includegraphics[width=0.18\textwidth]{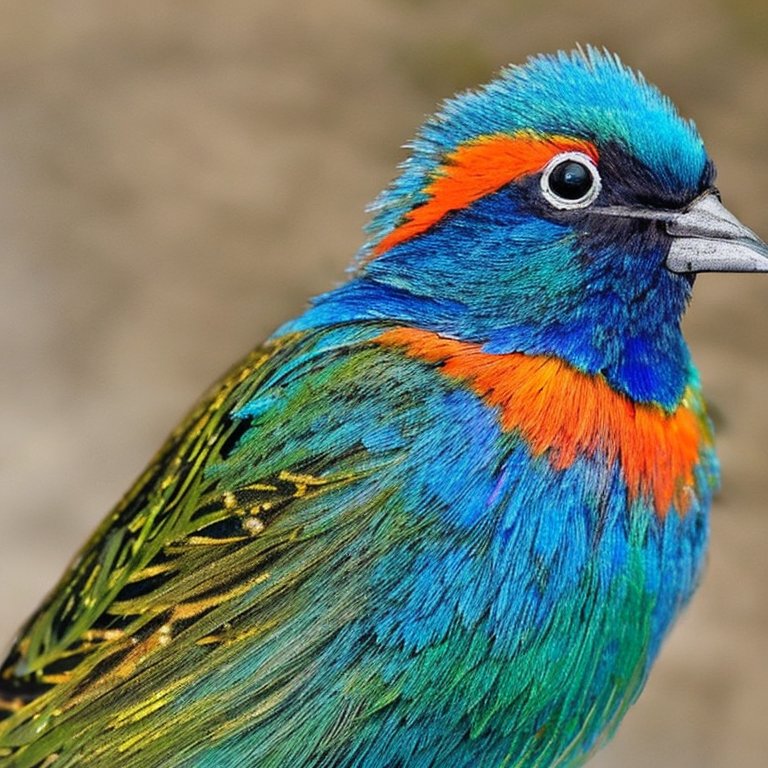} &
\includegraphics[width=0.18\textwidth]{pic/bird/Indigo_Bunting_0047_12966_f2.png} \\
\multicolumn{5}{p{1.0\textwidth}}{\textbf{caption:} a small colorful bird with teal feathers covering its body, with green speckles on its vent and abdomen.} \\ \hline

\shortstack{Class 014 \\ Indigo \\ Bunting \\ $0059\_11596.png$ \\ \qquad \\ \qquad \\ \qquad} &
\includegraphics[width=0.18\textwidth]{pic/bird/Indigo_Bunting_0059_11596.png} &
\includegraphics[width=0.18\textwidth]{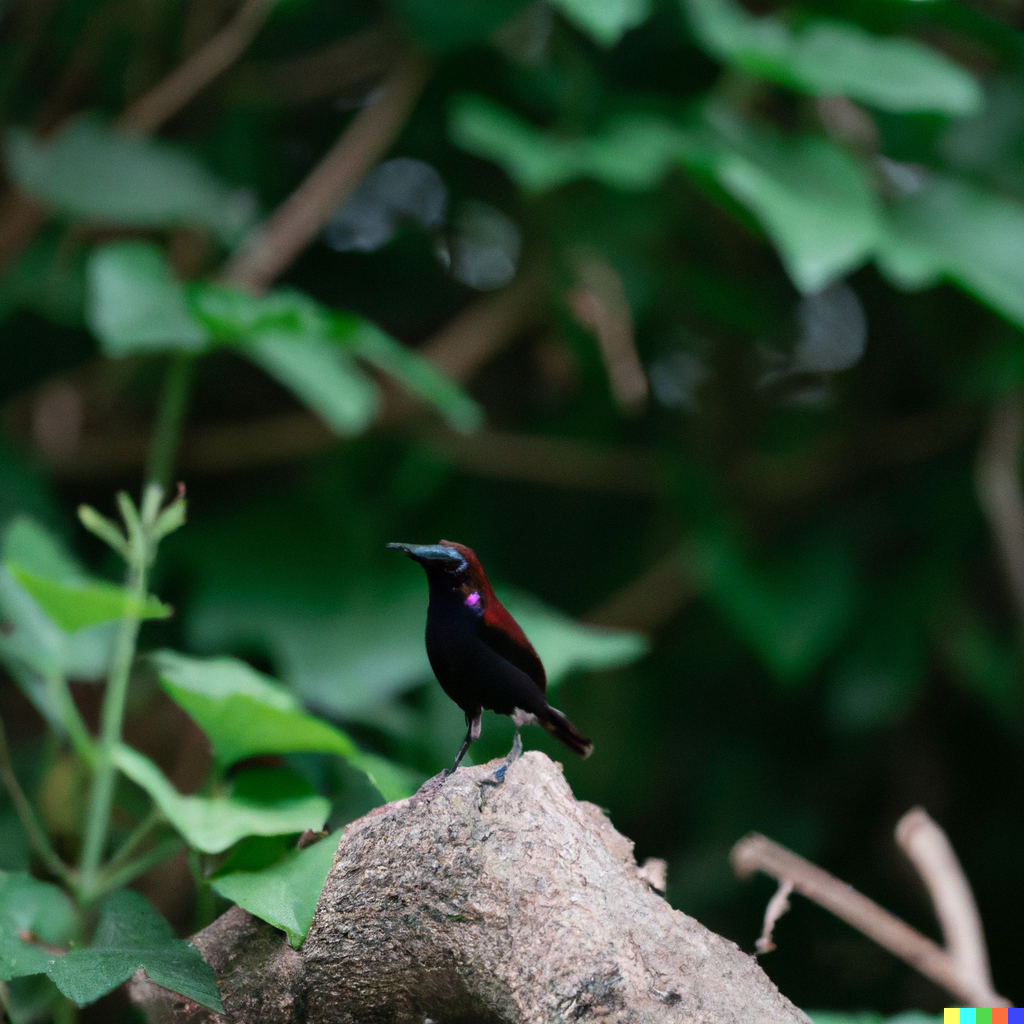} &
\includegraphics[width=0.18\textwidth]{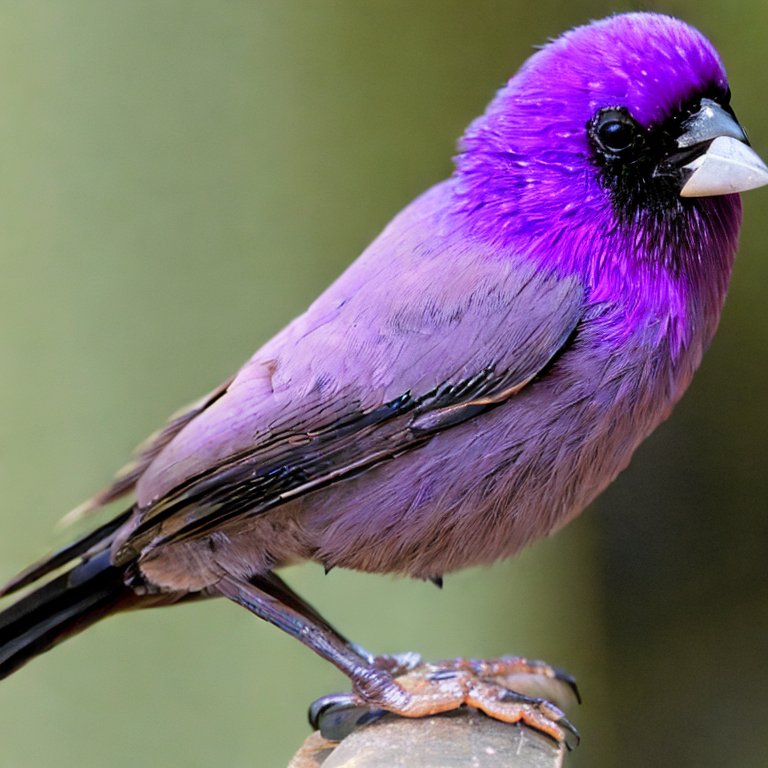} &
\includegraphics[width=0.18\textwidth]{pic/bird/Indigo_Bunting_0059_11596_f2.png} \\
\multicolumn{5}{p{1.0\textwidth}}{\textbf{caption:} a small purple bird, with black primaries, and a thick bill.} \\ \hline
\end{tabular}
\caption{Examples of generated images using DALLE-2, Stable Diffusion, and the proposed FG-RAT GAN trained on the CUB bird dataset. Each row represents a different sample (image size=256x256). The first column is the sample detail including class and specific image name. The second column is the corresponding target image. The third column is a generated image from DALLE-2. The fourth column is a generated image form Stable Diffusion. The fifth column is a generated image from our proposed FG-RAT GAN. As we can see, our proposed FG-RAT GAN can generate more realistic images where each image is similar to other images within the same class. For example, in the 1st row the proposed FG-RAT GAN generates a bird with dark brown body and white band encircling near the bill as specified in the caption, in the 3rd row it generates a bird with all gray body as specified in the caption, and both examples are similar to each other given that they belong to the same class.}
\label{fig: bird}
\end{figure}

\begin{figure}
\center{
\begin{tabular}{ccccc}
\hline
Class & Target & DALLE-2 & Stable Diffusion & FG-RAT GAN \\ \hline
\shortstack{Class 032 \\ $image\_05587.png$ \\ \qquad \\ \qquad \\ \qquad \\ \qquad \\ \qquad \\ \qquad} &
\resizebox{0.18\textwidth}{!}{\rotatebox{0}{\includegraphics{pic/flower/image_05587.png}}} &
\resizebox{0.18\textwidth}{!}{\rotatebox{0}{\includegraphics{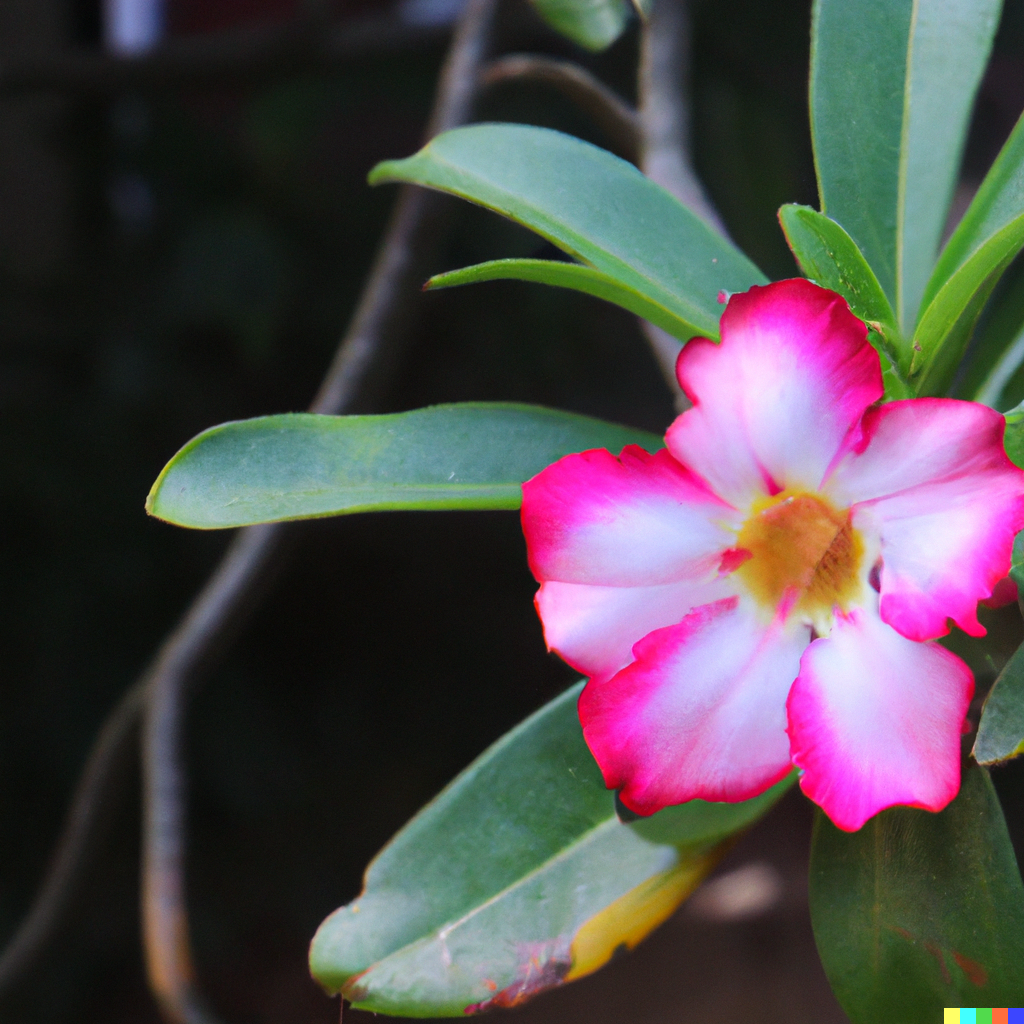}}} &
\resizebox{0.18\textwidth}{!}{\rotatebox{0}{\includegraphics{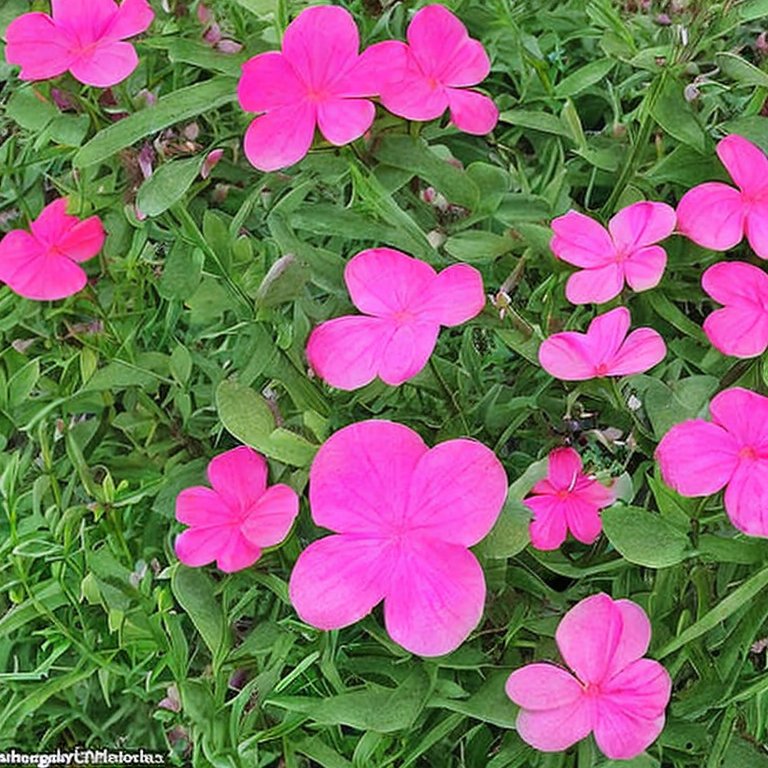}}} &
\resizebox{0.18\textwidth}{!}{\rotatebox{0}{\includegraphics{pic/flower/image_05587_f2.png}}} \\
\multicolumn{5}{p{1.0\textwidth}}{\textbf{caption:} the petals of the flowers are various shades of pink and have five individual petals.} \\ \hline

\shortstack{Class 032 \\ $image\_05602.png$ \\ \qquad \\ \qquad \\ \qquad \\ \qquad \\ \qquad \\ \qquad} &
\resizebox{0.18\textwidth}{!}{\rotatebox{0}{\includegraphics{pic/flower/image_05602.png}}} &
\resizebox{0.18\textwidth}{!}{\rotatebox{0}{\includegraphics{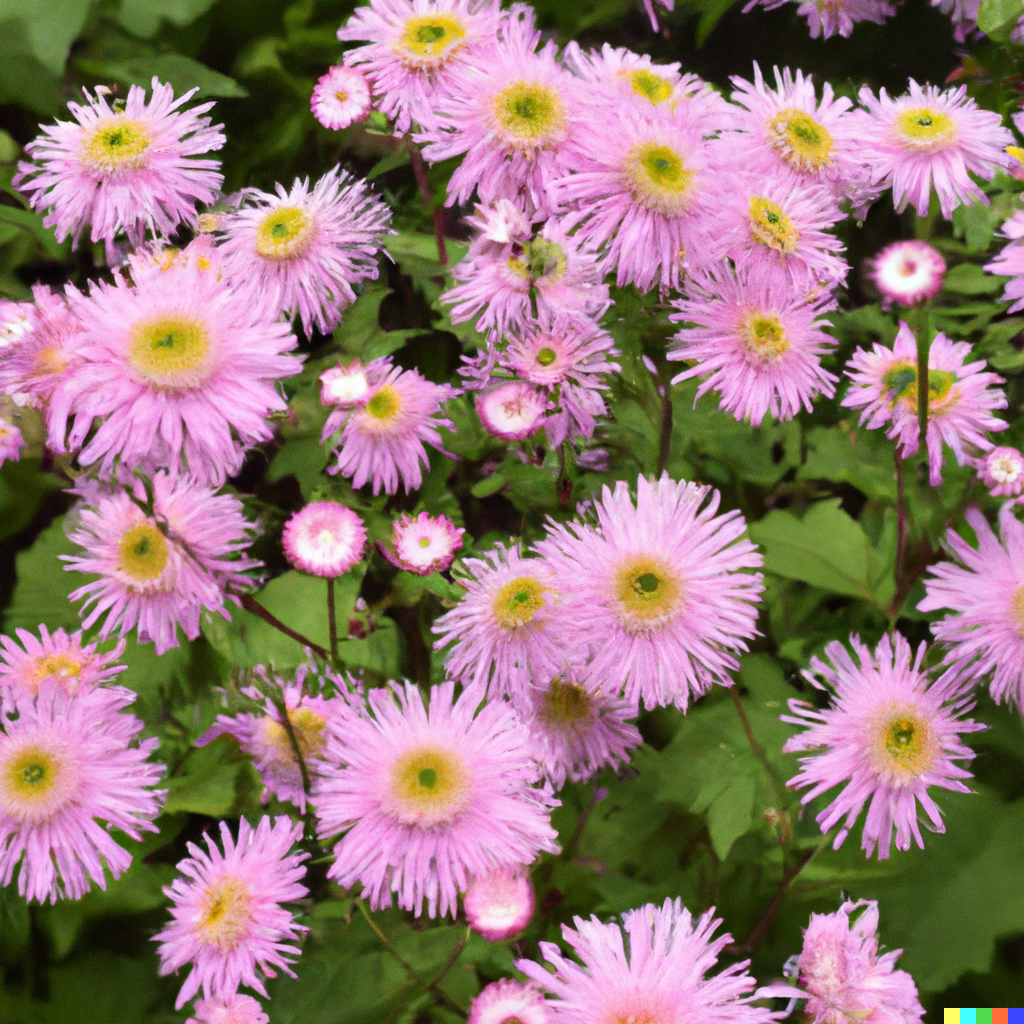}}} &
\resizebox{0.18\textwidth}{!}{\rotatebox{0}{\includegraphics{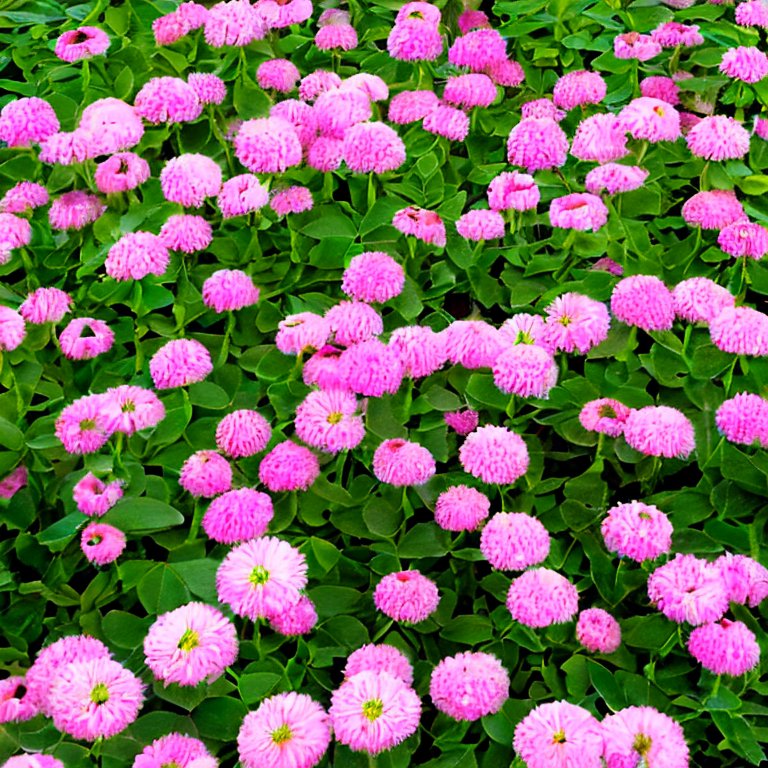}}} &
\resizebox{0.18\textwidth}{!}{\rotatebox{0}{\includegraphics{pic/flower/image_05602_f2.png}}} \\
\multicolumn{5}{p{1.0\textwidth}}{\textbf{caption:} a large group of light pink flowers with dark pink centers.} \\ \hline

\shortstack{Class 032 \\ $image\_05604.png$ \\ \qquad \\ \qquad \\ \qquad \\ \qquad \\ \qquad \\ \qquad} &
\resizebox{0.18\textwidth}{!}{\rotatebox{0}{\includegraphics{pic/flower/image_05604.png}}} &
\resizebox{0.18\textwidth}{!}{\rotatebox{0}{\includegraphics{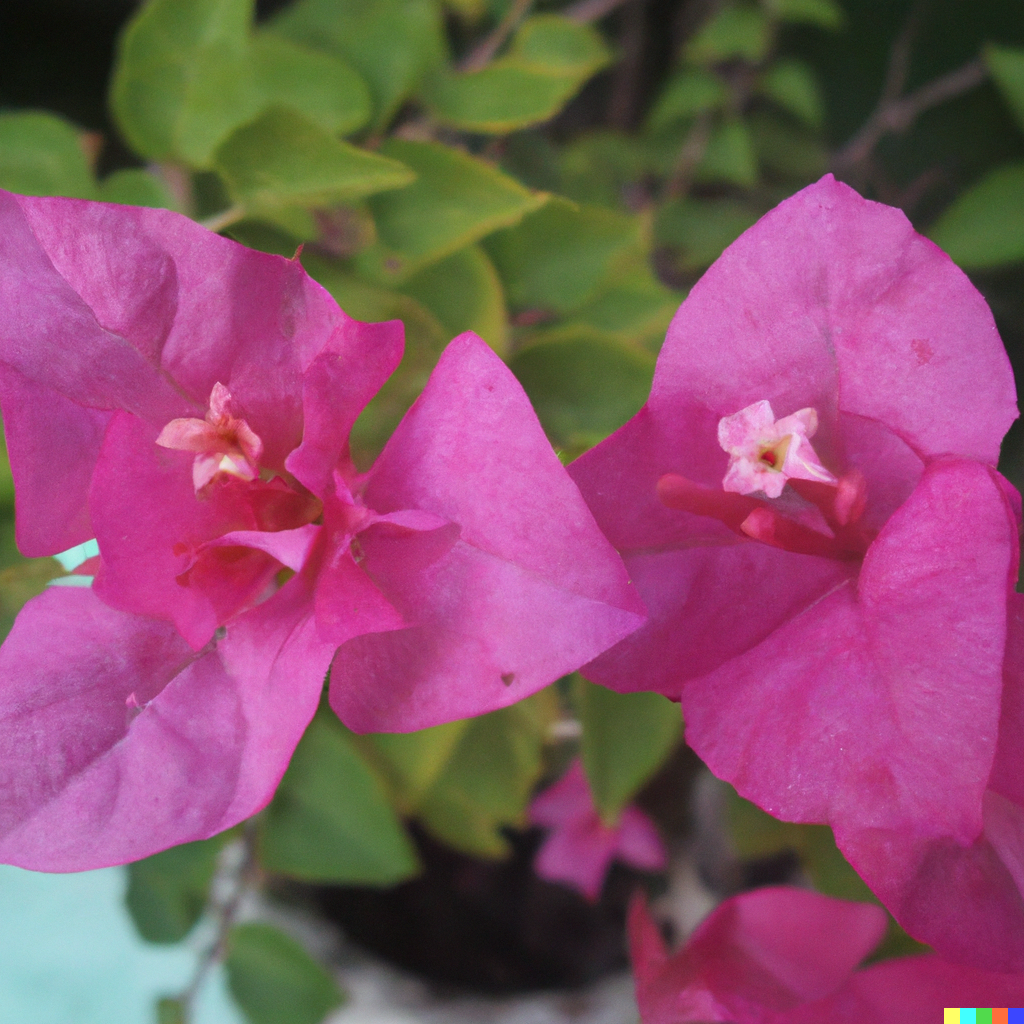}}} &
\resizebox{0.18\textwidth}{!}{\rotatebox{0}{\includegraphics{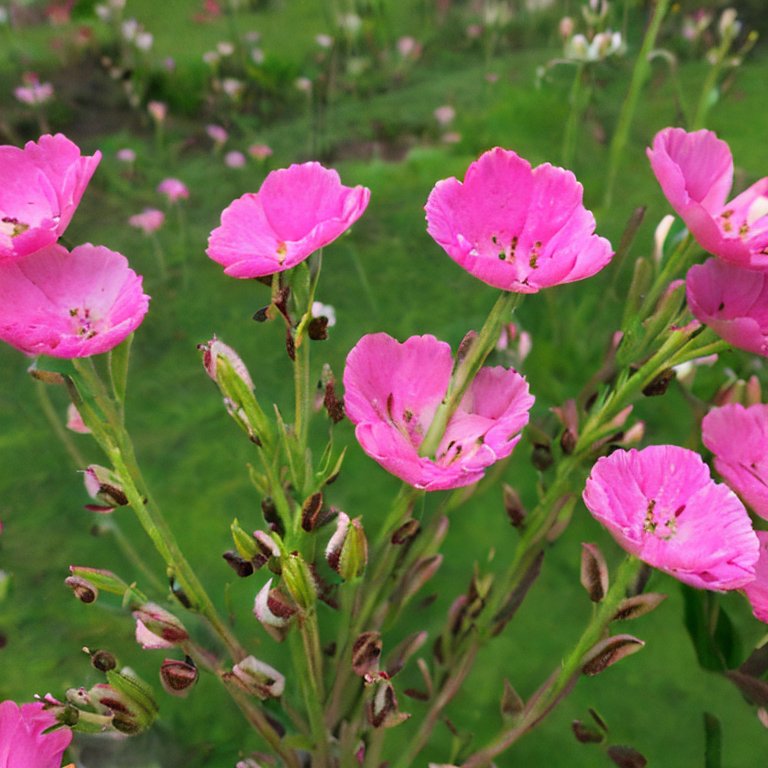}}} &
\resizebox{0.18\textwidth}{!}{\rotatebox{0}{\includegraphics{pic/flower/image_05604_f2.png}}} \\
\multicolumn{5}{p{1.0\textwidth}}{\textbf{caption:} these flowers are mostly pink but some of them have white parts located closer to their stamens.} \\ \hline

\shortstack{Class 049 \\ $image\_06209.png$ \\ \qquad \\ \qquad \\ \qquad} &
\resizebox{0.18\textwidth}{!}{\rotatebox{0}{\includegraphics{pic/flower/image_06209.png}}} &
\resizebox{0.18\textwidth}{!}{\rotatebox{0}{\includegraphics{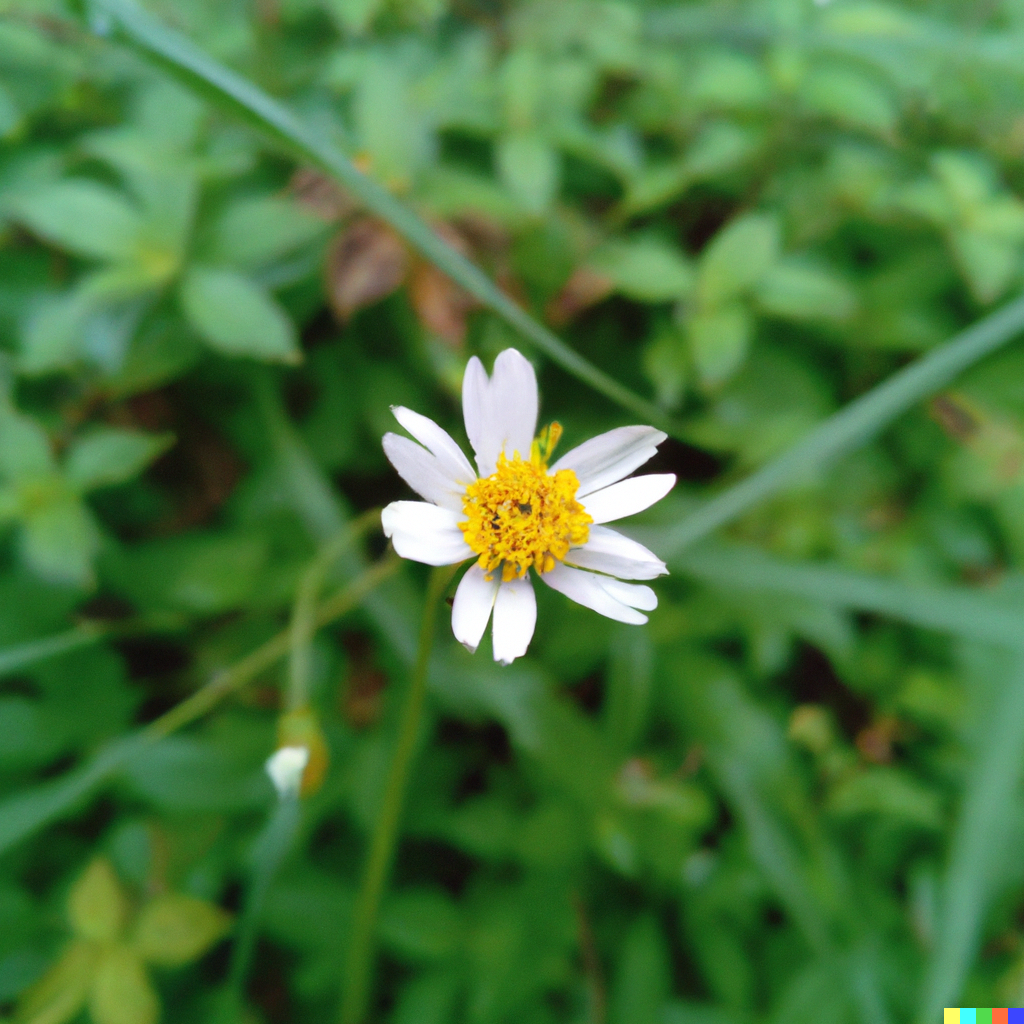}}} &
\resizebox{0.18\textwidth}{!}{\rotatebox{0}{\includegraphics{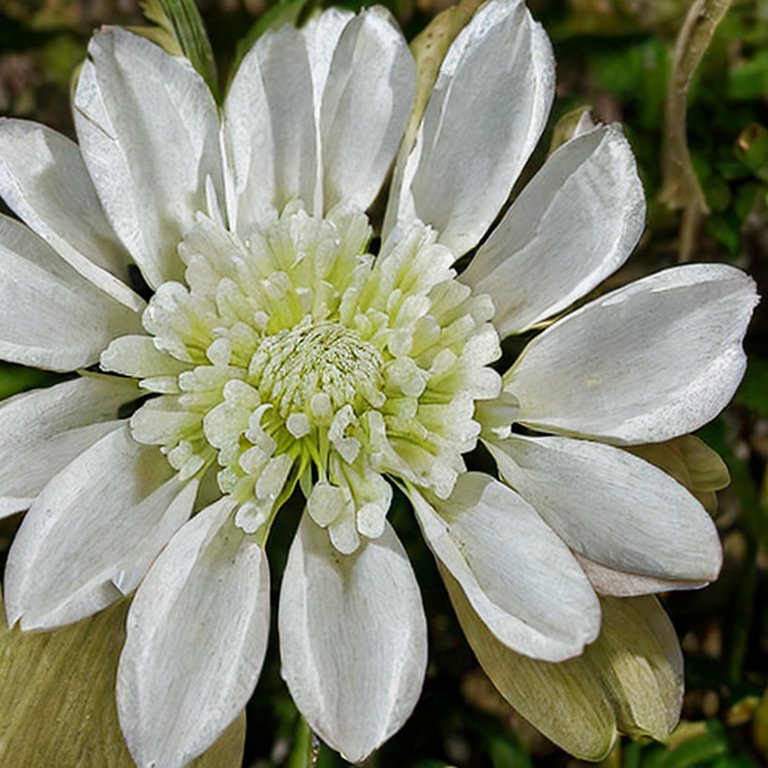}}} &
\resizebox{0.18\textwidth}{!}{\rotatebox{0}{\includegraphics{pic/flower/image_06209_f2.png}}} \\
\multicolumn{5}{p{1.0\textwidth}}{\textbf{caption:} this flower has thin white petals as its main feature.} \\ \hline

\shortstack{Class 049 \\ $image\_06216.png$ \\ \qquad \\ \qquad \\ \qquad} &
\resizebox{0.18\textwidth}{!}{\rotatebox{0}{\includegraphics{pic/flower/image_06216.png}}} &
\resizebox{0.18\textwidth}{!}{\rotatebox{0}{\includegraphics{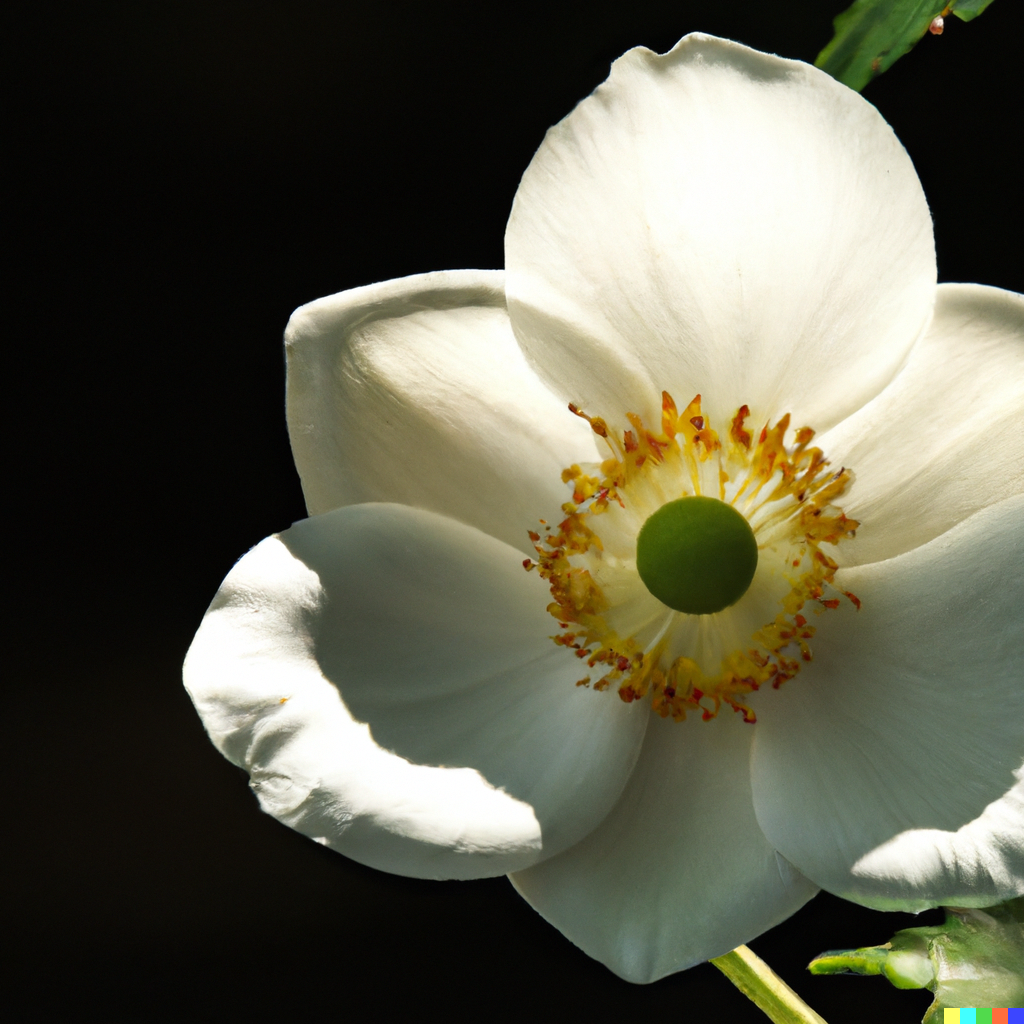}}} &
\resizebox{0.18\textwidth}{!}{\rotatebox{0}{\includegraphics{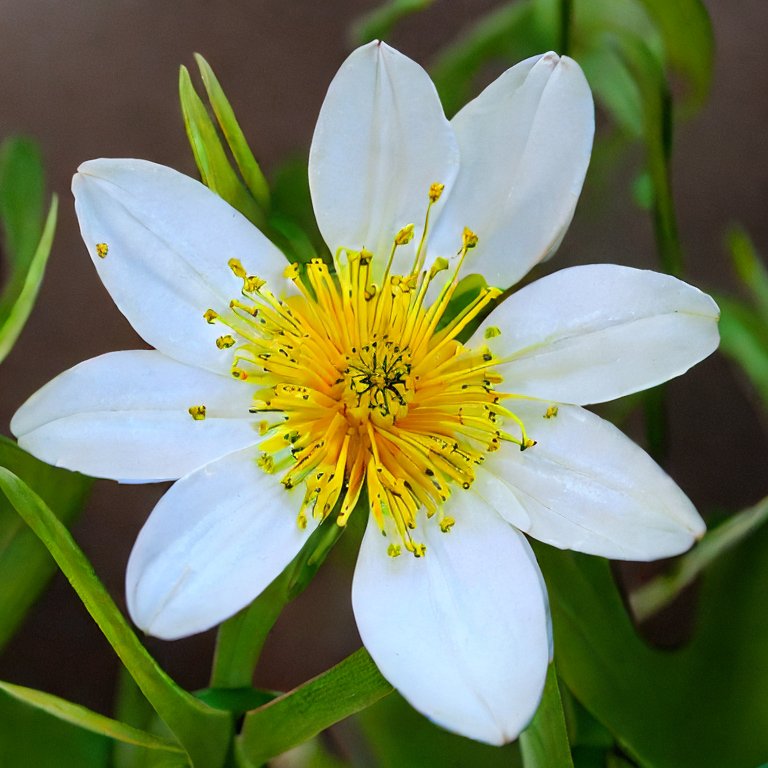}}} &
\resizebox{0.18\textwidth}{!}{\rotatebox{0}{\includegraphics{pic/flower/image_06216_f2.png}}} \\
\multicolumn{5}{p{1.0\textwidth}}{\textbf{caption:} the petals on this flower are white with yellow stamen.} \\ \hline

\shortstack{Class 049 \\ $image\_06224.png$ \\ \qquad \\ \qquad \\ \qquad} &
\resizebox{0.18\textwidth}{!}{\rotatebox{0}{\includegraphics{pic/flower/image_06224.png}}} &
\resizebox{0.18\textwidth}{!}{\rotatebox{0}{\includegraphics{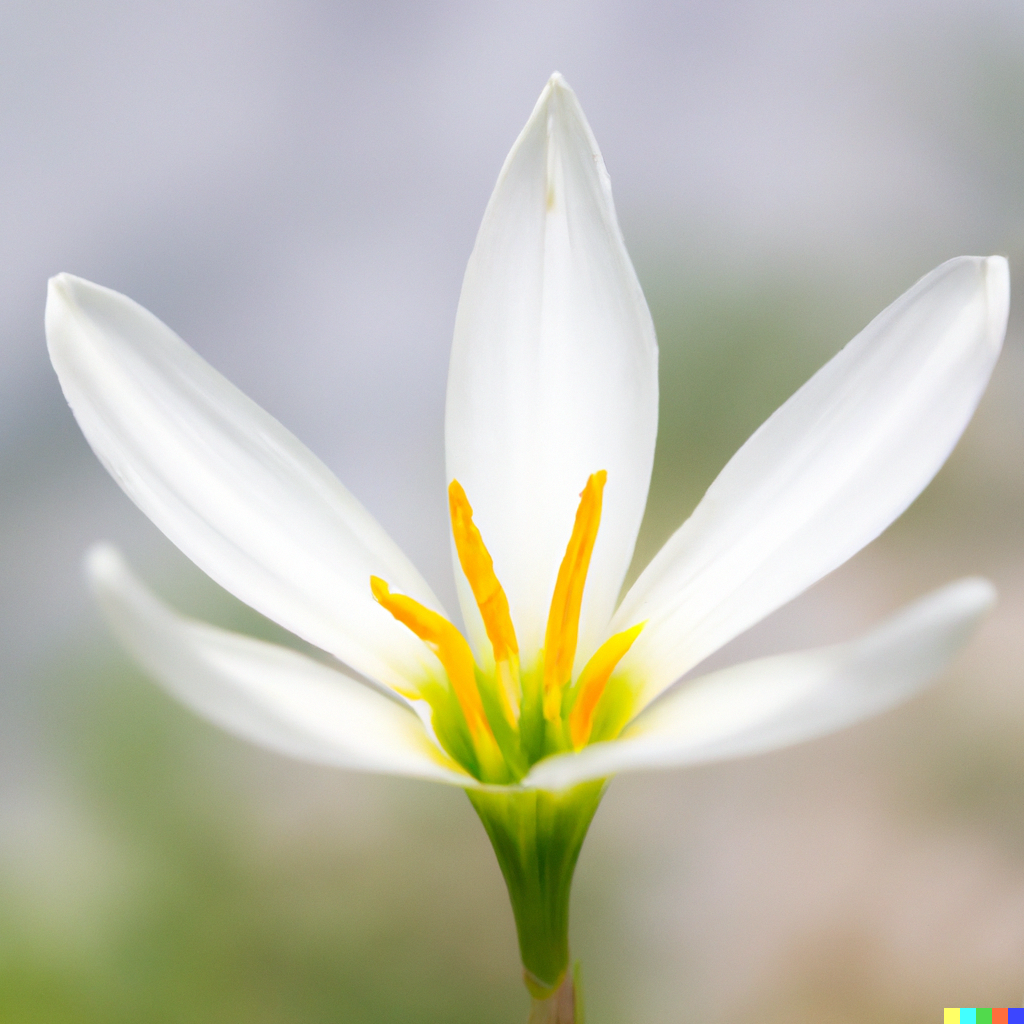}}} &
\resizebox{0.18\textwidth}{!}{\rotatebox{0}{\includegraphics{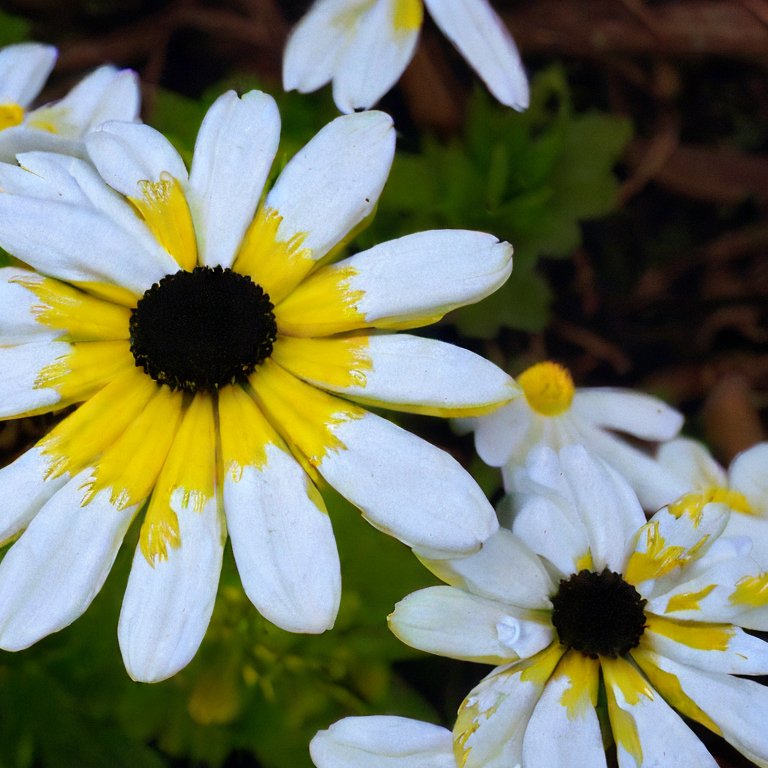}}} &
\resizebox{0.18\textwidth}{!}{\rotatebox{0}{\includegraphics{pic/flower/image_06224_f2.png}}} \\
\multicolumn{5}{p{1.0\textwidth}}{\textbf{caption:} the flower has petals of a white color with a many yellow stamen.} \\ \hline 
\end{tabular}}
\caption{Examples of generated images using DALLE-2, Stable Diffusion, and the proposed FG-RAT GAN trained on the Oxford flower dataset. Each row represents a different sample (image size=256x256). The first column is the sample detail including class and specific image name. The second column is the corresponding target image. The third column is a generated image from DALLE-2. The fourth column is a generated image form Stable Diffusion. The fifth column is a generated image from our proposed FG-RAT GAN. As we can see, our proposed FG-RAT GAN can generate more realistic images where each image is similar to other images within the same class. For example, the 5th row generates a flower with white petals and yellow stamen as in the description, the 6th row generates a flower with white petals and yellow stamen as in the description, and both samples are similar to each other given they belong to the same class.}
\label{fig: flower}
\end{figure}

\end{document}